%% file: ASG_CVPR.tex
\pdfoutput=1
\documentclass[10pt,twocolumn,letterpaper]{article}
\usepackage{cvpr}
\usepackage{float}
\include{formatAndDefs}

\cvprfinalcopy  
\def\cvprPaperID{10049} 

\begin{document}

\title{Appearance Shock Grammar for Fast Medial Axis Extraction from Real Images}
\author{
    Charles-Olivier Dufresne Camaro$^1$, Morteza Rezanejad$^4$,\\  
    Stavros Tsogkas$^{1,2}$\thanks{\scriptsize{Sven Dickinson and Stavros Tsogkas contributed to this article in their personal capacity as Professor and Adjunct Professor, respectively, at the University of Toronto. The views expressed (or the conclusions reached) are their own and do not necessarily represent the views of Samsung Research America, Inc.}}~, Kaleem Siddiqi$^4$, Sven Dickinson$^{1,2,3*}$
    \\[2mm]
    $^1$University of Toronto\ \
    $^2$Samsung Toronto AI Research Center\\
    $^3$Vector Institute for Artificial Intelligence\\ 
    $^4$School of Computer Science and Centre for Intelligent Machines, McGill University\\
   {\tt\small \{camaro,tsogkas,sven\}@cs.toronto.edu, \{morteza,siddiqi\}@cim.mcgill.ca}
}

\maketitle
\input{abstract}
\input{introduction}

\input{related}

\input{shocktheory}
\input{method}
\input{experiments}

\input{discussion}
\input{acknowledgements}
{   
    \small
    \bibliographystyle{ieee_fullname}
    \bibliography{ASG_CVPR}
}
\include{supplemental}

\end{document}

%% file: formatAndDefs.tex
\usepackage{times}
\usepackage[pdftex]{graphicx}
\usepackage{caption}
\usepackage{subcaption}
\usepackage{color}
\usepackage{xcolor}
\usepackage{amsmath,amssymb}
\usepackage[ruled, vlined, linesnumbered]{algorithm2e}
\usepackage{multirow}
\usepackage{array}
\usepackage{nameref}
\usepackage{xspace}
\usepackage{units}
\pdfoutput=1
\graphicspath{{./figures/}}
\DeclareGraphicsExtensions{.pdf,.jpeg,.png,.jpg}

\newcommand{\refsec}[1]{Section~\ref{#1}}
\newcommand{\reffig}[1]{Figure~\ref{#1}}
\newcommand{\refeq}[1]{Equation~\eqref{#1}}
\newcommand{\refalg}[1]{Algorithm~\ref{#1}}
\newcommand{\reftab}[1]{Table~\ref{#1}}

\renewcommand{\vec}[1]{\mathbf{#1}}
\newcommand{\set}[1]{\mathcal{#1}}

\newcommand{\R}{\mathbb{R}}

\def\x{\vec{x}}
\def\y{\vec{y}}
\newcommand{\f}[1]{\vec{f}_{#1}} 	 
\newcommand{\norm}[1]{\left \lVert #1 \right \rVert} 

\newcommand{\rad}[1]{r_{#1}} 

\makeatletter
\let\orgdescriptionlabel\descriptionlabel
\renewcommand*{\descriptionlabel}[1]{%
  \let\orglabel\label
  \let\label\@gobble
  \phantomsection
  \edef\@currentlabel{#1}%
  \let\label\orglabel
  \orgdescriptionlabel{#1}%
}
\makeatother

%% file: abstract.tex
\begin{abstract}
We combine ideas from shock graph theory with more recent appearance-based methods for medial axis extraction from complex natural scenes, improving upon the present best unsupervised method, in terms of efficiency and performance. 
We make the following specific contributions:
i) we extend the shock graph representation to the domain of real images, by generalizing the shock type
definitions using local, appearance-based criteria;
ii) we then use the rules of a Shock Grammar to guide our search for medial points, 
drastically reducing run time when compared to other methods, which exhaustively consider all points in the input image;
iii) we remove the need for typical post-processing steps including thinning, non-maximum suppression, 
and grouping, by adhering to the Shock Grammar rules while deriving the medial axis solution;
iv) finally, we raise some fundamental concerns with the evaluation scheme used in previous work and propose 
a more appropriate alternative for assessing the performance of medial axis extraction from scenes.
Our experiments on the BMAX500 and SK-LARGE datasets demonstrate the effectiveness of our approach. We outperform the present state-of-the-art, excelling particularly in the high-precision regime, while running an order of magnitude faster and requiring no post-processing.
\end{abstract}

%% file: introduction.tex
\section{Introduction}\label{sec:introduction}
\begin{figure*}[t!]
    \centering
    \includegraphics[width=\linewidth]{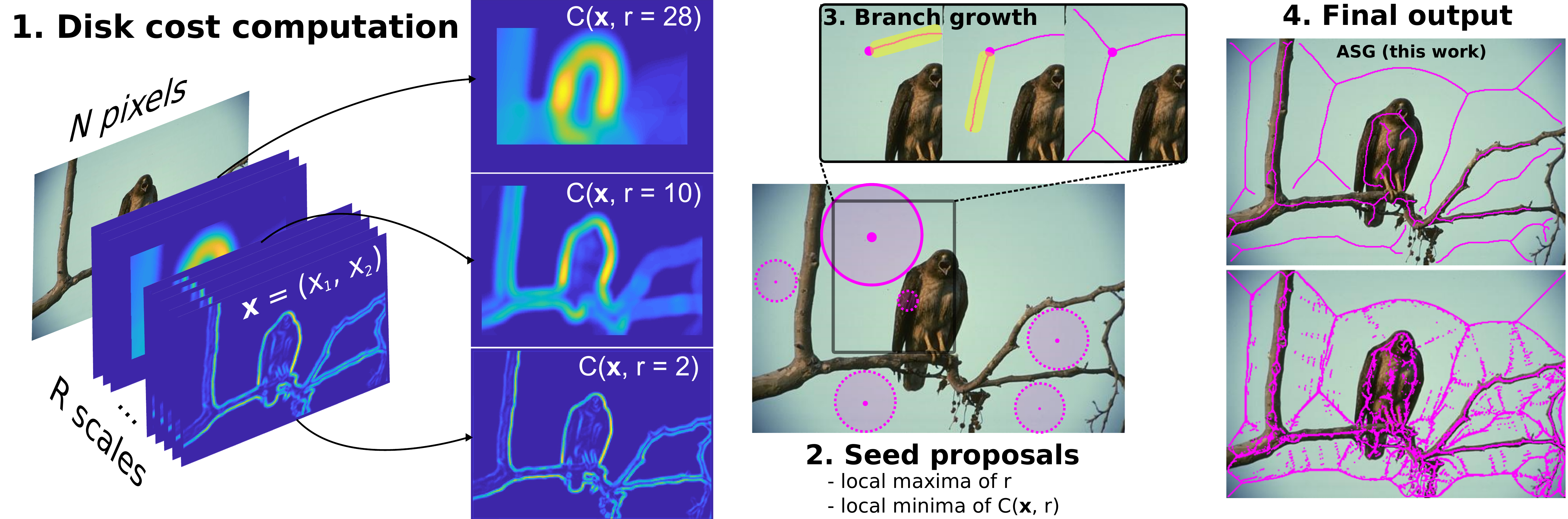}
    \caption{Our ASG algorithm consists of the following steps:
        \textbf{(1) Disk cost computation} associates each valid medial disk proposal with a cost $C(\x,\rad{})$. 
        Low costs (blue) represent high ``medialness'', whereas high costs (yellow) denote disks that span heterogeneous image regions.
        \textbf{(2) Seed proposals} are selected as 
        local \emph{scale maxima} and local \emph{disk cost minima} (example seeds and the respective disks are shown).
        \textbf{(3) Branch growth} of the selected seed into a medial branch.
        By following the rules of the SG grammar, the ASG only needs to 
        examine a small, fixed number of proposals in a scale-space neighborhood around a medial point  
        (shown in yellow), making it orders of magnitude faster than the AMAT~\cite{tsogkas2017amat},
        which naively considers $O(NR)$ medial disk proposals at each step.
        \textbf{(4) Final output} after growing all seeds. 
        The SG grammar rules automatically enforce connectivity and single-pixel 
        width constraints, producing a sparse, piecewise smooth scene medial axis.
        In comparison, the AMAT produces much noisier results, that require further post-processing.
    }
    \label{fig:teaser}
\end{figure*}
Object shape has a fundamental role in visual perception theory. 
Shape defines a basic level of abstraction that determines the spatial extent of 
structures in the physical world, and drives object recognition.
A popular representation of 2D shape is the \emph{Medial Axis Transform (MAT)}~\cite{blum1967transformation}.
The medial axis has been of particular interest in both human and 
computer vision because of its direct relationship to local symmetries of objects.
Local symmetries effectively decompose a shape into salient parts, aiding recognition and pose estimation, while being robust to viewpoint changes.
At the same time, symmetry in general has been proven to be instrumental in the 
analysis of complex scenes \cite{DBLP:journals/jvcir/SharvitCTK98,DBLP:conf/cvpr/TekSK97}, facilitating the encoding of shape and their 
discrimination and recall from 
memory~\cite{barlow1979versatility,royer1981detection,wagemans1998parallel}.
The importance of symmetry for scene categorization has been recently re-confirmed in \cite{rezanejad2018scene,wilder2019local}.

There are many algorithms that compute the MAT of 2D binary shapes. This problem was first discussed by 
Blum in his seminal
work~\cite{blum1967transformation,blum1973biological}, 
followed by several extensions and variants, including smooth local symmetries~\cite{brady1984smoothed}, 
shock graphs~\cite{sebastian2004recognition,siddiqi1999shock},
bone graphs~\cite{macrini2011bone,macrini2011object,macrini2008skeletons}
Hamilton-Jacobi skeletons~\cite{siddiqi2002hamilton}, 
augmented fast-marching~\cite{telea2002augmented},
hierarchical skeletons~\cite{telea2004optimal}, and the scale axis transform~\cite{giesen2009scale}. 

In an effort to broaden the application of such methods, interest in the problem of skeleton extraction 
from natural images has been recently revived, with a focus on using supervised learning.
The first such approach is that of Tsogkas and Kokkinos~\cite{tsogkas2012learning}, which was later followed by other methods, including the deployment of random
forests~\cite{teo2015detection}, or convolutional neural networks~\cite{funk2017symm,ke2017srn,Liu2017rsrn,Liu2017segskel,shen2017deepskeleton,wang2018deepflux,zhao2018hi}.
Departing from this trend, Tsogkas and Dickinson defined the first 
\emph{complete} MAT for color images, formulating medial axis 
extraction as a set cover problem~\cite{tsogkas2017amat}. 
However, all these recent approaches have an important limitation: medial points are extracted \emph{in isolation}, 
without explicit consideration of the local context, i.e., the structural 
constraints imposed by the fact that they must lie on skeletal segments within regions bounded by curves, with the associated generic classification of the medial axis point types \cite{DBLP:journals/ijcv/GiblinK03}.
As a result, one has to consider medial proposals at multiple
scales for each point, resulting in a very large space of medial point proposals to search. 
To make things worse, post-processing steps, such as non-maximum suppression, are required to ensure $1$-pixel width results, or to group medial points into meaningful segments.

In this paper we propose a method that reduces search space redundancies
and mitigates the need for post-processing in appearance-based medial point extraction 
from natural scenes, by combining ideas from shock graph (SG) theory~\cite{sebastian2002shock,DBLP:conf/cvpr/SiddiqiK96,siddiqi1999shock}
and the AMAT~\cite{tsogkas2017amat}.
Specifically, we use the notion of medial disk costs introduced 
in~\cite{tsogkas2017amat} to come up with new 
definitions for shock types, tailored to the domain of RGB images.
Our new shock type definitions have all the properties of their binary 
counterparts, unlocking the use of the SG grammar defined in~\cite{siddiqi1999shock}.
The grammar allows us to view medial point generation and grouping from natural images as a generative process, 
in the same spirit as the synthesis of binary shapes via a combination of birth, growth, and death rules, 
first proposed in \cite{DBLP:conf/cvpr/SiddiqiK96}. 
The explicit use of the grammar drastically reduces the size and complexity of the search space 
and imposes structural regularities in detection, improving performance and computational efficiency.
This is visually illustrated in~\reffig{fig:teaser}.

The benefits of our technique are noteworthy: 
our \emph{appearance shock grammar (ASG)} results in a $\mathbf{11\times}$ 
\textbf{speed-up}  with respect to the original AMAT algorithm, without the need for postprocessing;
the use of the ASG ensures that all 
resulting medial points are connected to form single pixel-wide medial branches. 
We also raise concerns with the standard evaluation benchmarks 
that involve multiple scene skeleton ground truths, 
such as \emph{SYMMAX300}~\cite{tsogkas2012learning} and \emph{BMAX500}~\cite{tsogkas2017amat}.
We propose an alternative evaluation protocol that addresses these issues, 
and also takes into account the relative importance of individual medial points with regard to boundary reconstruction. 
On this improved protocol, the ASG outperforms the state-of-the-art in \emph{unsupervised} medial axis 
extraction by $7.6\%$, with a sparser and piecewise smoother output.

%% file: related.tex
\section{Related Work} \label{sec:related}

\paragraph{Shock-based medial axis extraction.}
Blum defined medial axes as the loci of the centers of all disks that can be maximally inscribed in the interior of the shape~\cite{blum1967transformation,blum1973biological}.
An equivalent definition involves \emph{shocks}~\cite{lax1971shock}, the points where ``grassfire'' wavefronts, starting from the boundaries of the shape, meet.
Siddiqi \etal~\cite{DBLP:conf/cvpr/SiddiqiK96,siddiqi1999shock} assume that the shocks composing 
the medial axis of a bounding contour are first computed and then introduce the concept of \emph{shock graphs}.
They ``color'' shocks into different types according to the local variation of the medial axis radius function, and then define a shock grammar 
that determines how shocks of different types are connected with one another.
The shock grammar can be used to convert a skeleton into a directed acyclic graph, for use by graph matching algorithms to perform shape matching.
Shock graphs have also been successfully used in recognition~\cite{sebastian2004recognition} and database indexing~\cite{sebastian2002shock}.
\emph{Bone graphs}~\cite{macrini2011bone,macrini2011object,macrini2008skeletons} build on shock graphs by decomposing 
medial axes into object parts, leading to related graphs for object recognition applications. 
Medial branches corresponding to salient object parts are tagged as \emph{bones}, 
while \emph{ligature} segments \cite{August1999ligature} connect the bones together.

\paragraph{Medial axis extraction in natural images.}
Most recent work on skeleton extraction from natural images relies on 
supervised learning.
Tsogkas and Kokkinos~\cite{tsogkas2012learning} propose a multiple instance learning approach combined with hand-crafted features, tailored specifically to 
local reflective symmetries.
Teo \etal~\cite{teo2015detection} improve on this approach by using a 
more powerful random forest classifier, and by encouraging global symmetric consistency through an MRF representation.
Shen \etal~\cite{shen2017deepskeleton} introduced the first deep-learning approach to solving this problem, where a fully convolutional neural network 
(CNN) extracts the locations of the skeleton points, while estimating the local medial disk radii,
by combining deep features at multiple scales.
Ke \etal~\cite{ke2017srn} propose a similar framework that stacks Residual Units in its side outputs, 
improving performance and robustness.
In contrast to works that simply fuse (concatenate) side-output responses, 
Zhao \etal~\cite{zhao2018hi} create an explicit hierarchy of skeleton features at different scales.
This allows for the refinement of responses at finer scales using high-level semantic context, 
but also of coarser scale responses by using high-detail local responses from early layers of the CNN.
Finally, Wang \etal~\cite{wang2018deepflux} frame the skeleton extraction 
problem as a 2D vector field generation problem using a CNN, where each vector 
maps an image point to a skeleton point, similar to the Hamilton-Jacobi skeleton 
algorithm~\cite{dimitrov2003flux,siddiqi2002hamilton}. 

A completely different, unsupervised approach, the AMAT, was proposed by
Tsogkas and Dickinson~\cite{tsogkas2017amat}. 
The AMAT frames medial axis extraction in color images as a geometric set cover
problem and solves it using a greedy approximate solution~\cite{vazirani2013approximation}.
The cost assigned to each potential covering element (disk) is provided by a 
function that prioritizes the selection of maximal disks, leading to a solution
approximating the medial axes of structures in the scene.

In the present paper, we use the same concept of disk costs to 
generalize the definitions of shocks~\cite{siddiqi1999shock} 
and, in turn, exploit the shock graph theory in the RGB domain.
Unlike~\cite{siddiqi1999shock}, we do not assume the medial axis is given. 
Rather, we use the rules of the SG grammar to
constrain the number of eligible medial disks that are considered at every step.
This allows us to be much more efficient than the AMAT~\cite{tsogkas2017amat}, 
where disks at all possible locations and scales are valid candidates for the 
greedy algorithm.

%% file: shocktheory.tex
\section{Shock Theory} \label{sec:shocktheory}
A shock graph (SG)~\cite{DBLP:conf/cvpr/SiddiqiK96, siddiqi1999shock} is a directed acyclic graph (DAG) built from a skeleton. 
Its nodes correspond to connected components of shocks of the same type, and its edges represent connections between these components.
The direction of an edge indicates the direction of the medial axis radius derivative between the coarser scale and the finer scale shock. 
The root of the graph is called the \emph{birth shock}.

Shocks represent a colouring for medial points with specific scale (medial axis radius) gradients. 
A \textbf{type 4} shock (blob) corresponds to a single medial point that is a local maximum in scale. 
Its counterpart, the \textbf{type 2} shock (neck), represents a single medial point that is 
a local minimum in scale and splits its medial branch into separate parts when removed.
\textbf{Type 3} shocks (ribbons or bends) are sets of connected medial points of equal scales. 
Finally, \textbf{type 1} shocks (protrusion) are sets of connected medial points 
with monotonically decreasing scales in one direction.

Formally, the shocks can be defined as follows. 
For a given closed shape $X$, let $M(X)$ be its medial axis representation. $M(X)$ consists of medial points $\x$ of scales $R(\x) \equiv R_{\x}$.
For a medial point $\x \in M(X)$ and an open disk $D(\x,\epsilon)$ of radius $\epsilon$ centered at $\x$, let $N(\x,\epsilon) = M(X) \cap D(\x,\epsilon)\setminus \{\x\}$ represent its $\epsilon$-neighbourhood. $\x $ is
\begin{description}
    \item[type 4\label{eq:shock_type4}] if $\exists \epsilon > 0 \text{ s.t. } R_{\x} > R_{\y},\       \forall \y \in N(\x,\epsilon)$; 
    \item[type 3\label{eq:shock_type3}] if $\exists \epsilon > 0 \text{ s.t. }  R_{\x} = R_{\y},\     \forall \y \in N(\x,\epsilon) \neq \emptyset $;
    \item[type 2] if $\exists \epsilon>0 \text{ s.t. } R_{\x} < R_{\y},\
        \forall \y \in N(\x,\epsilon) \neq \emptyset $ \\
        and $N(\x,\epsilon)$ is not connected;   
    \item[type 1] otherwise.
\end{description}
While the shock graph represents the relations between connected medial points in terms of their radii, 
the shock graph grammar reverses the underlying grassfire flow in time.
The successive application of its rules defines a generative process that grows parts of an object.
The birth rule dictates that \emph{birth shocks can only be types 3 or 4}, while the death 
rules allow the shock graph to terminate at any shock type.
The protrusion rules define how an interval of medial points, with a monotonically changing radius value, can attach at junctions.
Finally, the union rules define the conditions under which distinct branches can be connected together.

\subsection{Defining Shocks for Natural Images} \label{sec:shocktheory:color}
    \begin{figure}
        \centering
        \begin{subfigure}[t]{0.25\textwidth}
            \includegraphics[width=0.90\textwidth]{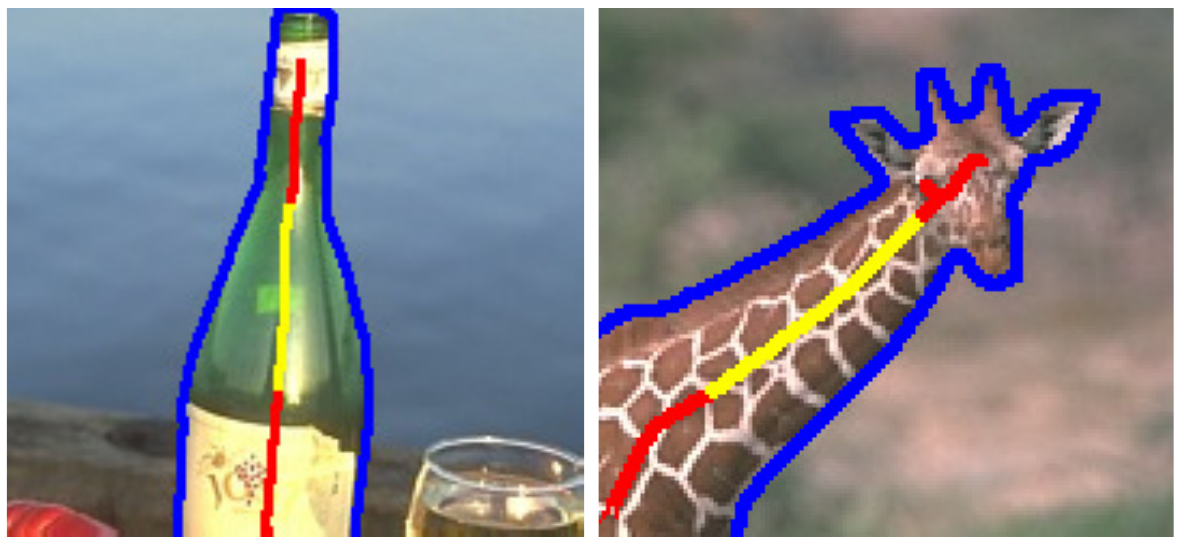}
            \caption{1-shocks (protrusion).}
            \label{fig:1shocks}
        \end{subfigure}%
        \begin{subfigure}[t]{0.25\textwidth}
            \includegraphics[width=0.95\textwidth]{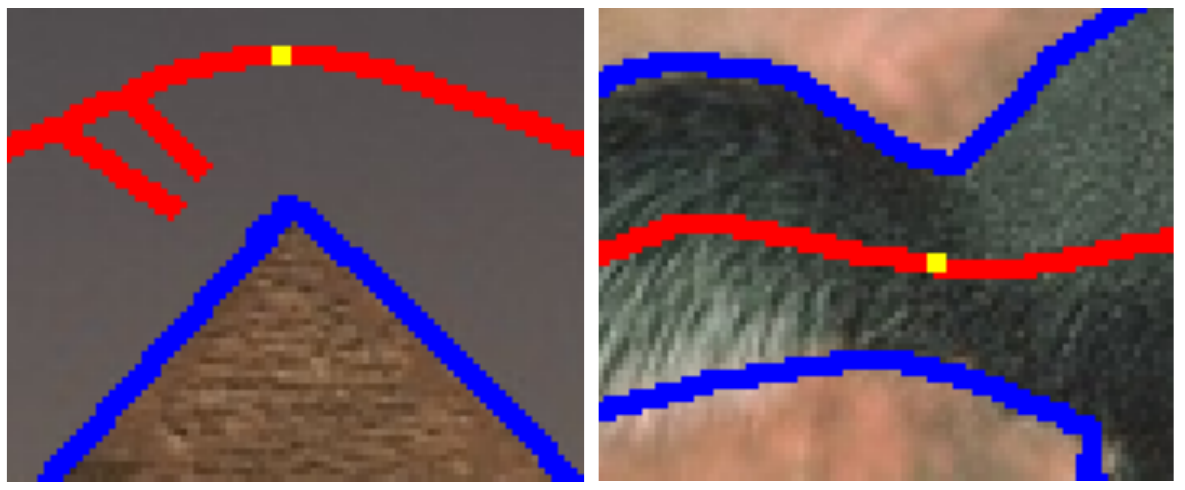}
            \caption{2-shocks (neck).}
            \label{fig:2shocks}
        \end{subfigure}
        \begin{subfigure}[t]{0.25\textwidth}
            \includegraphics[width=0.95\textwidth]{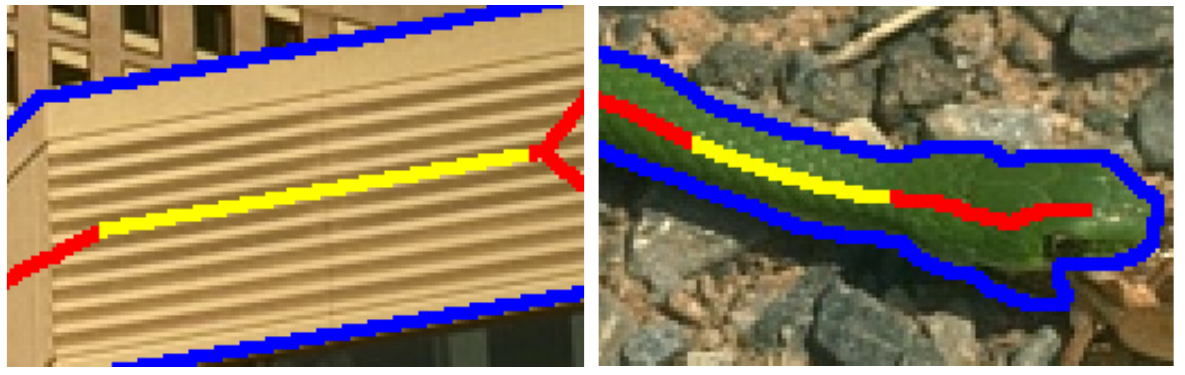}
            \caption{3-shocks (ribbon).}
            \label{fig:3shocks}
        \end{subfigure}%
        \begin{subfigure}[t]{0.25\textwidth}
            \includegraphics[width=0.9\textwidth]{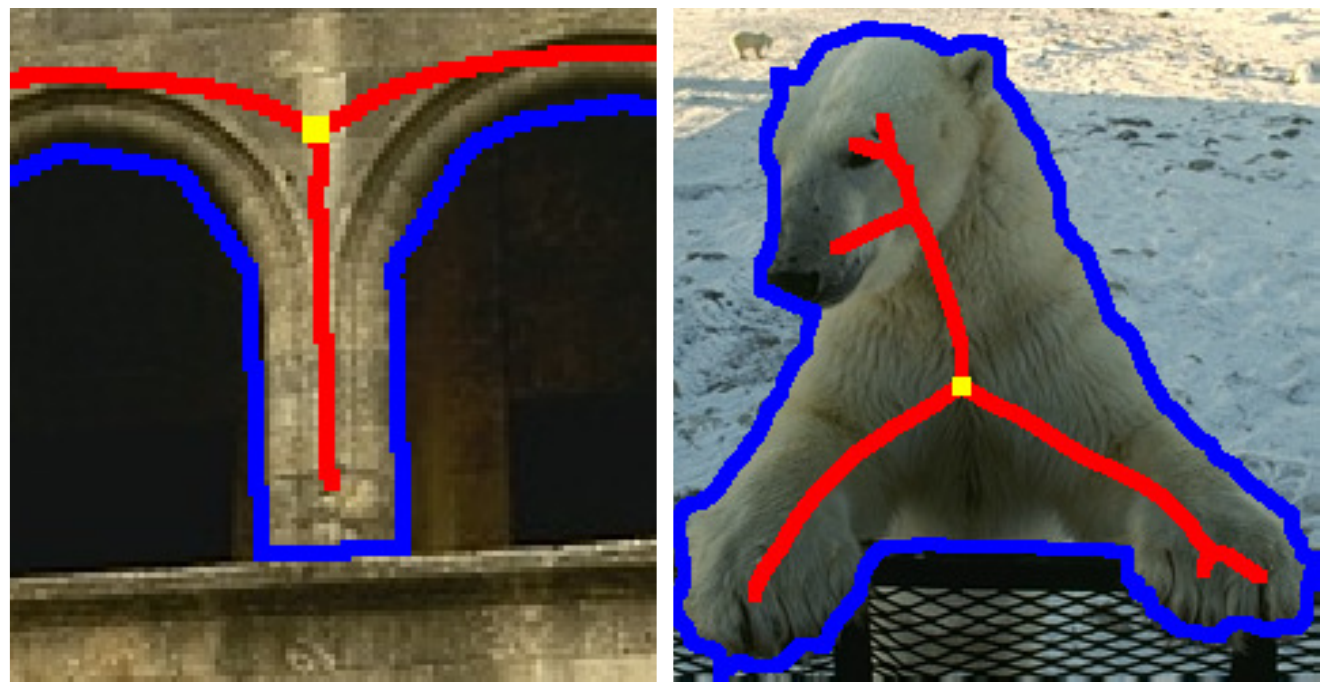}
            \caption{4-shocks (blob).}
            \label{fig:4shocks}
        \end{subfigure}
        \caption{Appearance based shock type examples from BMAX500 \cite{tsogkas2017amat}. 
            Medial axes are shown in red, contour in blue and selected shocks in yellow.
        } 
        \label{fig:rgb_shocks}
    \end{figure}    
    The ideas presented in~\refsec{sec:shocktheory} assume that $M(X)$
    has already been extracted using some skeletonization algorithm.
    In this work we turn the problem on its head: rather than using the shock grammar 
    to define a graph on a pre-existing skeleton, we use the rules imposed by the 
    grammar to constrain the search space for medial points.
    To do that, first we have to formally extend the shock type definitions to the domain
    of natural images.
    We employ the same notation as in~\refsec{sec:shocktheory}, introducing new notation when needed.

    The key component for determining the coloring of a shock in the binary domain is the
    computation of $R(\x)$, the radius of the largest disk, centered at $\x$, that remains
    contained in the open interior $\set{X}$ of a closed 2D shape. The contour of such disks
    is tangent to the shape's boundary at 2 points, at least.
    Exact computation of $R(\x)$ is feasible because the boundary of a 2D shape is 
    well defined (i.e., the points where the image values change from ``0'' to ``1''
    or vice-versa).
    This is not the case in the natural image domain, where extracting object boundaries
    is an ill-posed problem that typically admits a probabilistic solution. 
    
    To deal with this ambiguity, we follow the region-based approach of~\cite{tsogkas2017amat}
    and assign a cost $C(\x,r)$ to each disk proposal $D(\x, r) = D_{\x,\rad{}}$.
    This cost acts as a ``soft maximality'' indicator:
    if $\rad{}$ is close to the ideal (maximal) value, $C(\x,\rad{})$ is low, whereas disks that are 
    not maximal or cross image boundaries, are severely penalized. 
   
    More concretely,
    let $\x \in \R^2, \y \in N(\x, \epsilon)$ be medial points, and $R_{\x}, R_{\y} \in \R$ denote 
    the radii of the respective \emph{maximal} disks centered at $\x$ and $\y$.
    Also, let a small quantity $\delta_r > 0$ denote an acceptable ``cost margin'' for determining disk maximality,
    and $\epsilon_r > 0$.
    Intuitively, if $C(\x, \rad{}+\epsilon_r) - C(\x, \rad{}) < \delta_r$, then 
    $D_{\x, \rad{}+\epsilon_r}$ is a better candidate for being the maximal disk centered at $\x$
    than $D_{\x, \rad{}}$.
    We formalize the \emph{scale maximality criterion} as follows:
    \begin{equation}
        \label{eq:maximality}
        C(\x, R_{\x}) + \delta_r < C(\x, R_{\x}+\epsilon_r).
    \end{equation}
    This condition should be satisfied for all disk proposals that are added to our solution.
    We also define a ``cost smoothness'' criterion, expressing the fact that the costs 
    of neighboring medial points should not vary significantly.
    This is another direct analogy to the shock theory for binary shapes, 
    which dictates that the \emph{radii} of neighboring medial points are bound to vary slowly. 
    This is due to the fact that 
    shocks coincide with singularities of a  continuous Euclidean distance function from the boundary 
    \cite{blum1973biological}.
    Letting $\delta_c>0$,  we define the cost \emph{smoothness criterion} as
    \begin{equation}
        \label{eq:smoothness}
        \norm{C(\x, R_{\x}) - C(\y, R_{\y})} < \delta_c.
    \end{equation}
    By combining these two criteria with the binary shock type definitions,
    we redefine shock coloring rules in the RGB domain.
    These rules are agnostic to the exact nature of the cost function -- we discuss potential choices for $C$ in~\refsec{sec:method:implementation}. 
    Note, however, that, contrary to the binary case, we must consider all possible locations and scale candidates $(\x,\rad{\x})$, since we know neither the centers $\x \in M(X)$ 
    nor the radii $R$ of the true medial disks.
    Finally, our shock coloring definitions are adapted to accommodate a discrete pixel grid.
    For instance, the neighbourhood of a point $N(\x,1)$ corresponds to its immediate 
    8-connected neighbours, while radii only take positive integer values.

%% file: method.tex
\section{Constrained Medial Point Search Using the Shock Graph Grammar} \label{sec:method}
    The formal definition for RGB shocks described in~\refsec{sec:shocktheory:color}
    allows one to use the SG grammar to progressively build an object skeleton 
    while constraining the search space of candidate medial points.
    We summarize the steps of such an approach in \refalg{alg:overview}. 
    \begin{algorithm}
        \caption{Overview of algorithm}
        \label{alg:overview}
        
        \KwIn{RGB image $I$}
        \KwOut{Medial points $M$}
        Initialization: $M \leftarrow \emptyset$\\
        $D \leftarrow$ generateProposals($I$);\\
        $Q_s$ $\leftarrow$ extractSeeds($D$);\\
        \While{ notEmpty($Q_s$)}{
        $(\x_s,\rad{s})$ $\leftarrow$ selectSeed($Q_s$);\\
        $M \leftarrow$ growSeed($(\x_s,\rad{s})$,$D$);\\
        $Q_s$ $\leftarrow$ pruneSeeds($Q_s$,$M$);\\
        }
        $M \leftarrow$ growEndPoints($D$,$M$):\\
    \end{algorithm}
    
    First, we generate medial disk (point) proposals $D_{\x,\rad{}}$ at multiple scales $\rad{}$.
    Second, we extract birth seeds $(\x_s,\rad{s})$~
    from the pool of proposals and store them in a queue $Q_s$. 
    We grow each seed into a medial axis, by iteratively attaching low-cost medial 
    points\footnote{In fact, we add medial fragments rather than individual points. 
    See the paragraph on ``Seed growth'' in this Section.}.
    Every time we attach a new point to the axis, we make sure this attachment is consistent with the 
    rules of the SG grammar, and that the medial axis remains connected and one-pixel wide. 
    We greedily continue growing an axis until no points can be added without violating one of these constraints,
    and then pick the next seed in $Q_s$ to grow.
    Note that, because birth seeds can only be type 3 or 4 shocks, which correspond to 
    local scale maxima, the medial axis is constructed in a coarse-to-fine manner.
    After $Q_s$ has been exhausted, we relax $\delta_c$ and grow branch end points that may have been cut short due to the cost constraint.
    This step allows the algorithm to extend branch growths into more expensive/ambiguous image regions for completeness.
    We now describe each one of these steps in more detail.
    
    \paragraph{Proposal generation.}\label{sec:method:proposals}
    Each medial disk candidate $ D_{\x,\rad{}}$ is associated with a cost $C(\x,\rad{})$
    that represents how close $D_{\x,\rad{}}$ is to being ``maximal''.
    In the domain of real images, a low value for $C$ is equivalent to a perceptually homogeneous 
    appearance within the disk-shaped region $D^I_{\x,\rad{}} \subset I$.
    In \refsec{sec:method:implementation} we describe in detail two options for $C$ based on: 
    i) RGB encodings~\cite{tsogkas2017amat}; and
    ii) image intensity histograms.
    We compute $C(\x,\rad{})$ for all points $\x$ in 
    the image, at all potential scales $\rad{} \in [\rad{min},\rad{max}]$.
    Proposals corresponding to disks that are not fully enclosed in the image are ignored. 

    \paragraph{Seed extraction} should only return type 3 or type 4 shocks.
    To extract 4-shock \emph{seed} candidates, we scan the space of positions and scales, 
    and check whether the \ref{eq:shock_type4} criterion holds.
    For 3-shock \emph{seed} candidates, we check if there is at least one valid neighbour 
    sharing the same scale, as per the shock \ref{eq:shock_type3}  definition.
    Finally, we impose an additional requirement: a type 3/4 shock $\x_s$ qualifies as a seed
    iff it corresponds to a \emph{local cost minimum}, i.e.,
    \begin{equation}
        \label{eq:minimum}
        C(\x_s, r_{\x_s}) \leq C(\y, r_{\y}), \ \forall \y \in N(\x_s, 1).
    \end{equation}
    All seed candidates are added into a queue $Q_s$.
    Because a 4-seed can eventually grow into a nearby 3-seed as the medial axis is formed
    (provided that both seeds are part of the same object),
    once a seed has stopped growing, we also remove any other seeds in $Q_s$ that have 
    been added to $M$.  
    
    \paragraph{Seed selection} follows a coarse-to-fine strategy. 
    We prioritize the selection of seeds with larger radii and lower costs $C$ because 
    we expect their cost computation to be less sensitive to noise, 
    resulting in more robust axis growth.

    \paragraph{Seed growth} involves attaching medial point proposals to a selected seed
    $(\x_s,\rad{s})$
    , following the shock grammar. 
    At each step, the least expensive valid proposal $(\x,\rad{\x})$
    ~in the neighborhood of the axis is added to $M$. 
    Proposals whose regions $D^I_{\x,\rad{}}$ are subsumed by $M^I$, 
    the union of disk regions centered at points in $M$, are ignored, as they offer no new information about the object's shape.
    The growth process ends when no more valid proposals can be added. 
    To emulate the cost constraint in the RGB shock coloring definitions, we introduce a cost upper bound
    \begin{equation}
        C_{tol} = C(\x_s,\rad{s})(1+\alpha_c) > 0,
    \end{equation}
    where $\alpha_c$ is a small arbitrary positive constant. 
    We ignore proposals with costs larger than $C_{tol}$ to ensure that the quality of attached points 
    does not degrade during growth.
    
    Single points do not provide sufficient spatial context for determining 
    robust axis growth directions. 
    To resolve these ambiguities we grow a seed by attaching \emph{fragments} of valid 
    connected medial points, $F$, instead.
    For simplicity, we model medial axis fragments $F$ as linear segments of length $l_F \leq l_{max}$, 
    producing a piecewise-linear approximation of the true medial axis. 
    To rank the quality of candidate fragments  we define a \emph{fragment cost}
    \begin{equation}
        \Bar{C}_F =\frac{\alpha(l_F)}{l_F}\sum_{j=1}^{l_F}C(\x_j,\rad{j}),
        \label{eq:fragment_cost}
    \end{equation}
    which is proportional to the mean cost of its constituent points.
    The more expensive a fragment, the less likely it is to be part of the medial axis. 
    To prioritize longer fragments, which provide more context, 
    $\Bar{C}_F$ is weighted by a length-dependent parameter $\alpha(l_F)$, i.e., 
    between two fragments with equal mean cost, the longer one will be selected.
    
    At each iteration, we generate multiple candidate fragments and add the 
    one with the lowest $\bar{C}_F$  to $M$.
    Growth then continues from the endpoint of the last added fragment. 
    This step is repeated until no more valid fragments, i.e., 
    fragments that follow the SG grammar and whose regions are not subsumed by $M^I$, 
    can be attached to the current medial branch. 
    In practice this can happen either because the branch is fully grown or 
    because the remaining fragment candidates are too expensive.
    Then, additional medial branches can be grown from the seed $(\x_s,\rad{s})$.
    
    A medial branch may also terminate at a \emph{junction point}: 
    a medial point from which multiple branches emerge\footnote{Formally, a junction point 
    is a medial point with at least 3 neighbours.}.
    In this case, new branches can also be grown from that point, as shown in 
    Figure~\ref{fig:4shocks}.
    To identify a junction point, we check if multiple fragments can be attached 
    to it without violating the SG grammar's protrusion rules.

    \paragraph{End point growth.} \label{sec:method:endpoint}
    Restricting the growth of medial branches using a cost-based threshold for medial fragments 
    promotes robustness and avoids committing to potentially erroneous growth paths.
    However, the resulting medial axes may not be fully fleshed out: 
    branches corresponding to the fine image details are grown last,
    and do not always survive this pruning step.
    To recover these lost medial branches, we perform a final refinement step: we revisit each medial \emph{end point} 
    and allow it to grow further by relaxing the tolerance constraint $C_{tol}$, 
    thus allowing less salient fragments to be added.
    The algorithm terminates when no more valid fragments can be added to any medial end point.

\subsection{Cost functions} \label{sec:method:implementation}

    \paragraph{Color homogeneity.}
    We use the default cost function $C$ of the AMAT~\cite{tsogkas2017amat}, 
    after smoothing the input image $I$ using~\cite{xu2011image}.
    The cost of a disk region $D^I_{\x,\rad{}}$ with area $A_{\rad{}}$ is
    \begin{equation}
        \label{eq:amat_cost_effective}
        C_{color}(\x,\rad{}) = \frac{c(\x,\rad{})}{A_{\rad{}}} + \frac{w_s}{\rad{}},
    \end{equation}
    where $c(\x,\rad{})$ represents a measure of homogeneity  based on $\f{\x,\rad{}}$, 
    the average CIELAB space value within $D^I_{\x,\rad{}}$ :
    \begin{equation}
        c(\x,\rad{})\hspace{-.15em} = \hspace{-.25em} \sum_{k}\hspace{-0.25em}\sum_{l}||\f{\x,\rad{}}-\f{\x_k,\rad{l}}||^2_2
        \  \forall k,l: D^I_{\x_k,\rad{l}}\hspace{-0.25em}\subset\hspace{-0.25em} D^I_{\x,\rad{}}.
        \label{eq:amat_cost}
    \end{equation}
    
    \paragraph{Intensity histogram.}
    While straightforward to compute,~\refeq{eq:amat_cost_effective} is sensitive to gradual changes in intensity. 
    We consider a more powerful cost function that is based on local histograms of image intensity and is
    more appropriate for applications to regions with texture.
    We first smooth the image using~\cite{xu2011image}. 
    Then, we precompute a tiling of the image using $6\times6$ squares. 
    For each tile we compute an average intensity value per color channel. 
    We then construct a local histogram $\mathbf{H}$ for each channel, by placing these averages into one
    of 10 bins.
    To compute $c(\x,\rad{})$, we replace the $l^2$~-norm in \refeq{eq:amat_cost} with a standard Bhattacharyya distance between normalized histograms $\mathbf{H_1},\mathbf{H_2}$
    \begin{equation}
        d_{\textbf{Bhatt.}}(\mathbf{H_1,H_2}) =\sqrt{1\hspace{-0.25em}-\hspace{-0.25em}\frac{\sum_i\sqrt{\mathbf{H_1}(i)\cdot
        \mathbf{H_2}(i)}}{\sqrt{\sum_i\mathbf{H_1}(i)} \cdot \sqrt{\sum_i\mathbf{H_2}(i)}}},
    \end{equation}
    averaged over the 3 color channels, as used for unsupervised texture segmentation in region-based active contours \cite{Michailovich07imagesegmentation}. 
    For each disk under consideration, the histograms are computed using only the enclosed tiles. 
    We also rescale, and add a scale-dependent constant 
    to obtain 
    \begin{equation}
        \label{eq:st_cost}
        C_{hist}(\x,\rad{}) = \frac{c(\x,\rad{})}{\rad{}} + \frac{w_s}{\rad{}}.
    \end{equation}
    

%% file: experiments.tex
\section{Experiments}\label{sec:experiments}

\begin{table*}
    \resizebox{\linewidth}{!}{
    \centering
    \begin{tabular}{c|c c|c c||c c|c c||c c|c c}
       Resolution & \multicolumn{8}{c||}{$161\times241$ pixels (half)} & \multicolumn{4}{c}{$321\times481$ pixels (full)} \\
        \hline
        Method ($C(\x,r)$) 
        & \multicolumn{2}{c|}{AMAT (Color)} & \multicolumn{2}{c||}{ASG (Color)}
        & \multicolumn{2}{c|}{AMAT (Hist)} & \multicolumn{2}{c||}{ASG (Hist)}  
        & \multicolumn{2}{c|}{AMAT (Hist)} & \multicolumn{2}{c}{ASG (Hist)} \\
        \hline 
        
        P 
        & .393 & {\color{blue} .237} & \bf{.396} & {\color{blue} \bf{.246}} 
        & .431 & {\color{blue} .268} & \bf{.506} & {\color{blue} \bf{.343}} 
        & .471  & {\color{blue} .295} & \bf{.641} & {\color{blue} \bf{.474}}\\ 
        
        R 
        & \bf{.640} & {\color{blue} \bf{.665}} & .452 & {\color{blue} .485} 
        & \bf{.623} & {\color{blue} \bf{.658}}  & .541 & {\color{blue} .595} 
        & \bf{.769} & {\color{blue} \bf{.794}} & .503 & {\color{blue} .546} \\ 
        
        F1 
        & \bf{.487} & {\color{blue} \bf{.350}} & .422 & {\color{blue} .326} 
        & .509 & {\color{blue} .380} & \bf{.522} & {\color{blue} \bf{.435}} 
        & \bf{.584} & {\color{blue} .431} & .564 & {\color{blue} \bf{.507}} \\ 
        \hline
        
        *R gains
        & \bf{+.043} & {\color{blue} \bf{+.047}} & +.032 & {\color{blue} +.039}  
        & +.016 & {\color{blue}+.020} & \bf{+.035} & {\color{blue}\bf{+.040}} 
        & +.018 & {\color{blue}+.020} & \bf{+.036} & {\color{blue}\bf{+.040}}\\ 
        
        *F1 gains 
        & +.012 & {\color{blue} +.006} & \bf{+.014} & {\color{blue}\bf{+.008}} 
        & +.006 & {\color{blue}+.004} & \bf{+.016} & {\color{blue}\bf{+.011}} 
        & +.005 & {\color{blue}+.003} & \bf{+.022} & {\color{blue}\bf{+.015}}\\\hline 
        
        t (s) 
        & \multicolumn{2}{c|}{57.4} & \multicolumn{2}{c||}{\bf{7.0}  ($\mathbf{\downarrow8.2\times}$)} 
        & \multicolumn{2}{c|}{33.7} & \multicolumn{2}{c||}{\bf{6.5}  ($\mathbf{\downarrow5.2\times}$)}  
        & \multicolumn{2}{c|}{393.2} & \multicolumn{2}{c}{\bf{34.7} ($\mathbf{\downarrow11.3\times}$)} \\ 
    \end{tabular}
    }
    \caption{
        Results on the BMAX500 using standard evaluation (black) and our proposed {\color{blue}single-annotation protocol (blue)}.
        Gains for the ligature-weighted version of BMAX500 are denoted by *.
        Timings are averages over the BMAX500 test set. 
        Cost function computation times are excluded from the runtime measurements to compare the two algorithms head to head.
    }
    \label{tab:bmaxhalfthin}
\end{table*}

We conduct experiments on scene and object skeleton detection, on two representative datasets:
\emph{BMAX500}~\cite{tsogkas2017amat} and \emph{SK-LARGE}~\cite{shen2017deepskeleton}.
BMAX500 is built by automatically extracting skeletons of human-annotated region segments from the 
BSDS500 dataset~\cite{amfm_pami2011}; each image typically comes with 5-7 such annotations.
We use the downsampled version of BMAX500 as in~\cite{tsogkas2017amat}, but 
we also evaluate on the full resolution dataset, 
to more effectively highlight the computational gains of our approach.
SK-LARGE, on the other hand, focuses on \emph{object-centric} skeleton detection: 
each image contains a centered object and the ground truth is only the \emph{foreground} 
object skeleton. 
Note that this is a different problem than the one the ASG (and the AMAT) aim to solve, making
comparison on the SK-LARGE unfair to our algorithm, but we still include it for completeness.

\subsection{Evaluation Protocol and Criticisms} \label{sec:experiments:evaluation}
\begin{figure}
    \centering
    \includegraphics[width=0.32\linewidth]{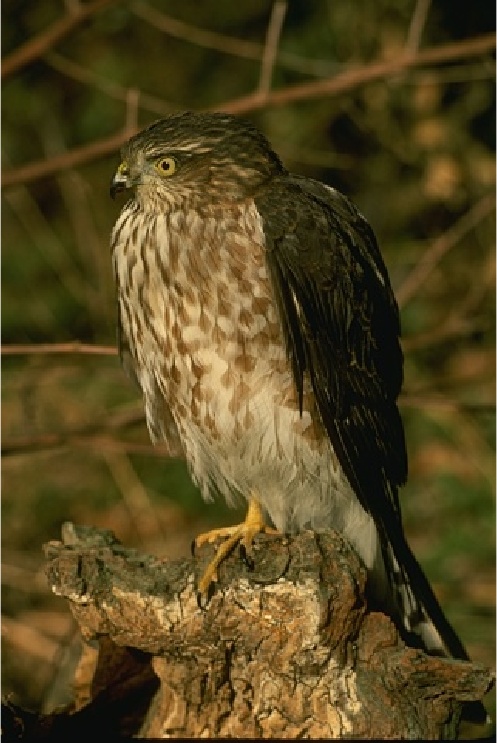}
    \includegraphics[width=0.32\linewidth]{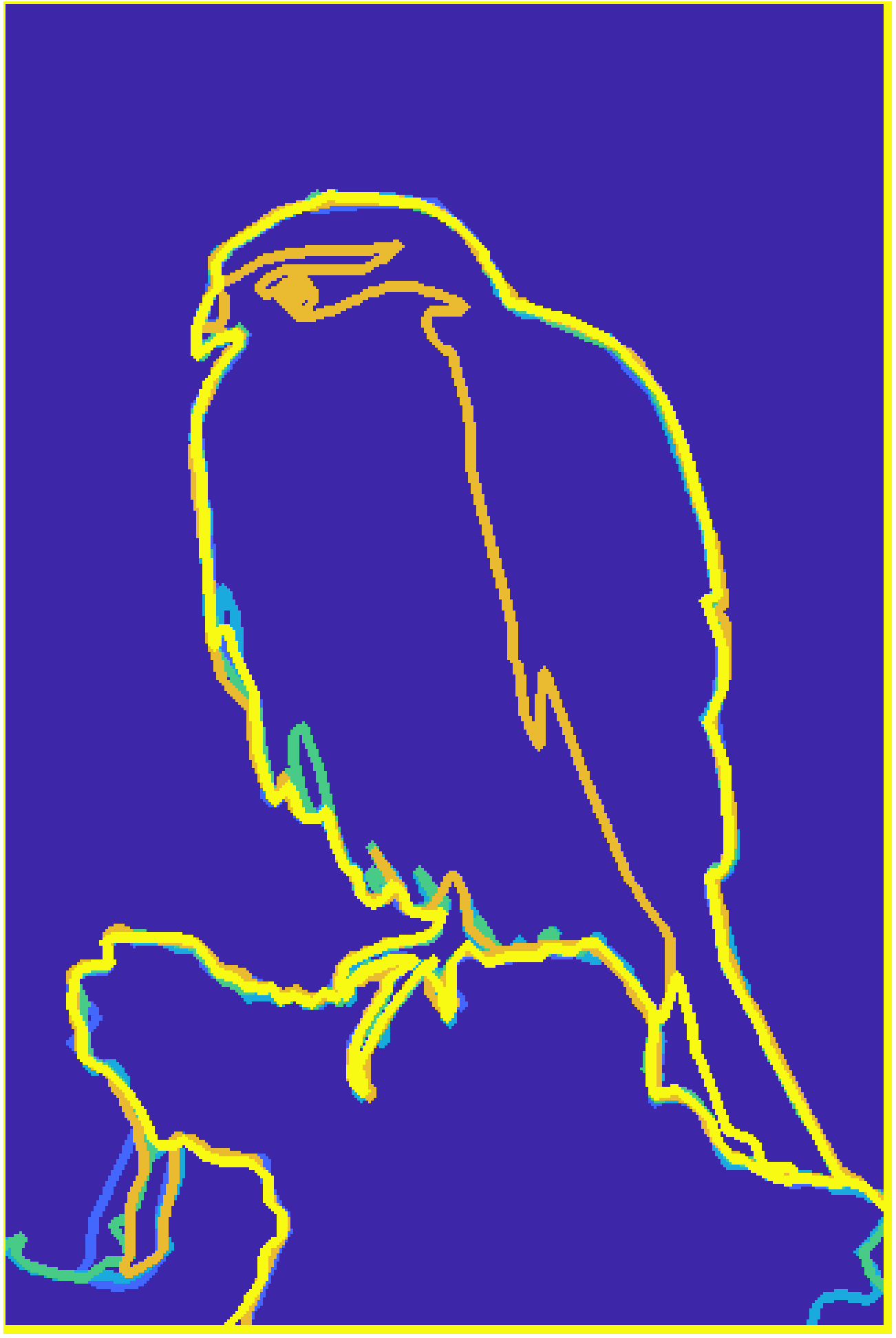}
    \includegraphics[width=0.32\linewidth]{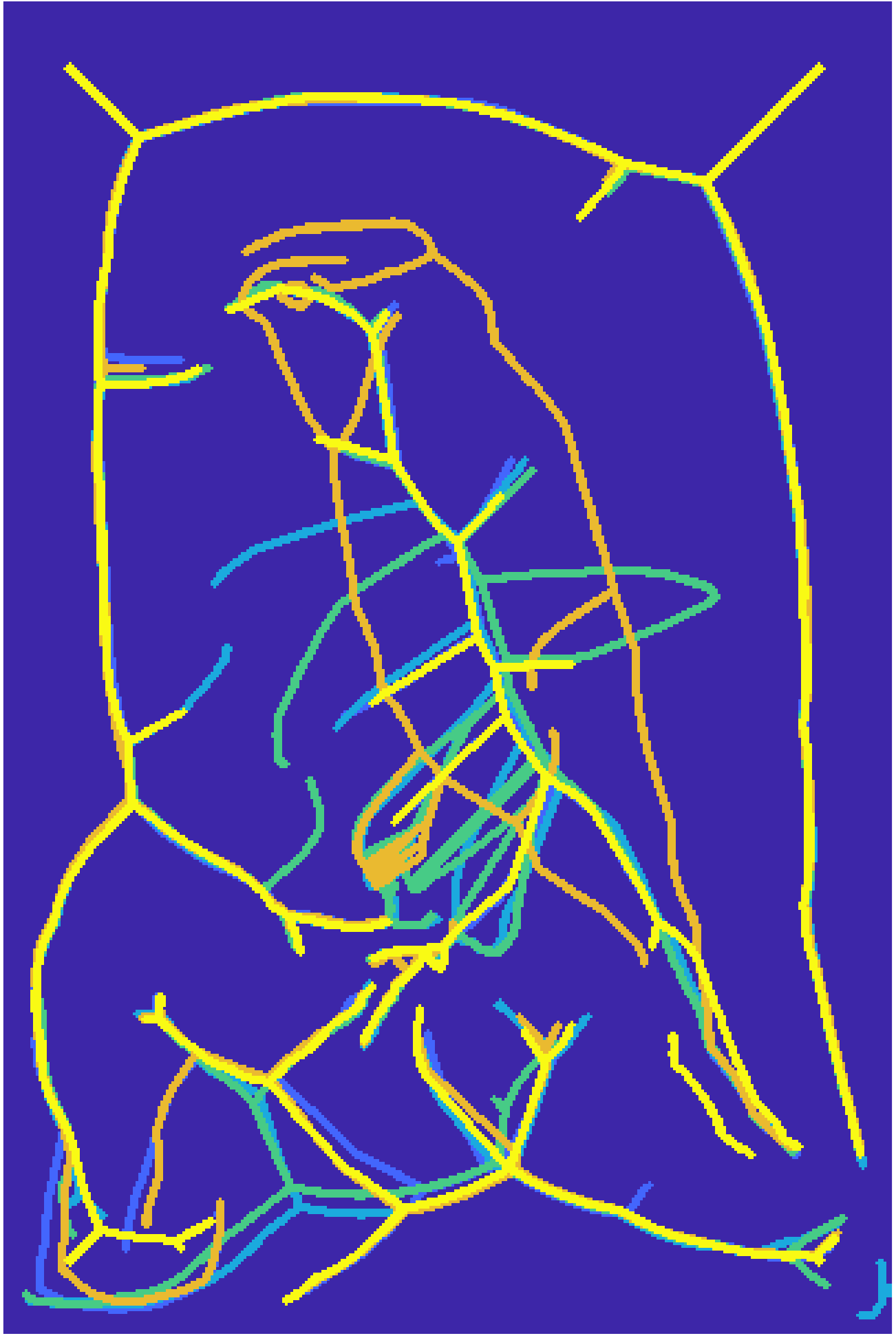}
    \caption{Boundary (middle) and skeleton (right) annotations on the BSDS/BMAX500.
        Different colors denote annotations extracted from different segmentations.
        Whereas boundaries for the same scene form a natural hierarchy,
        skeletons actually \emph{conflict} with one another, making the evaluation
        protocol used in~\cite{tsogkas2017amat} unsuitable.
    }
    \label{fig:boundary_vs_skeleton_gt}
\end{figure}

\begin{figure}
    \centering
    \newcommand\wfactor{0.325}
    \includegraphics[width=\wfactor\linewidth]{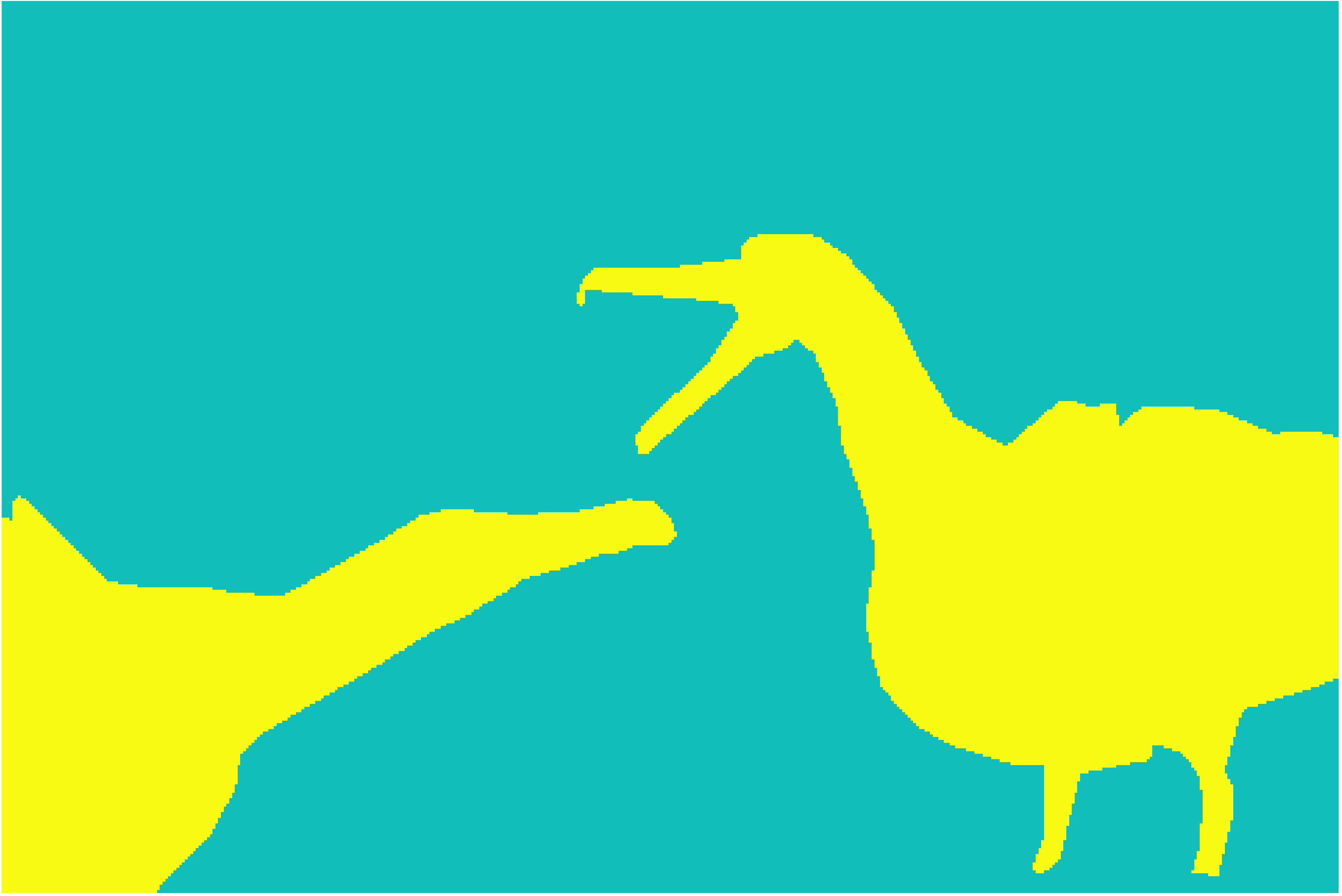}
    \includegraphics[width=\wfactor\linewidth]{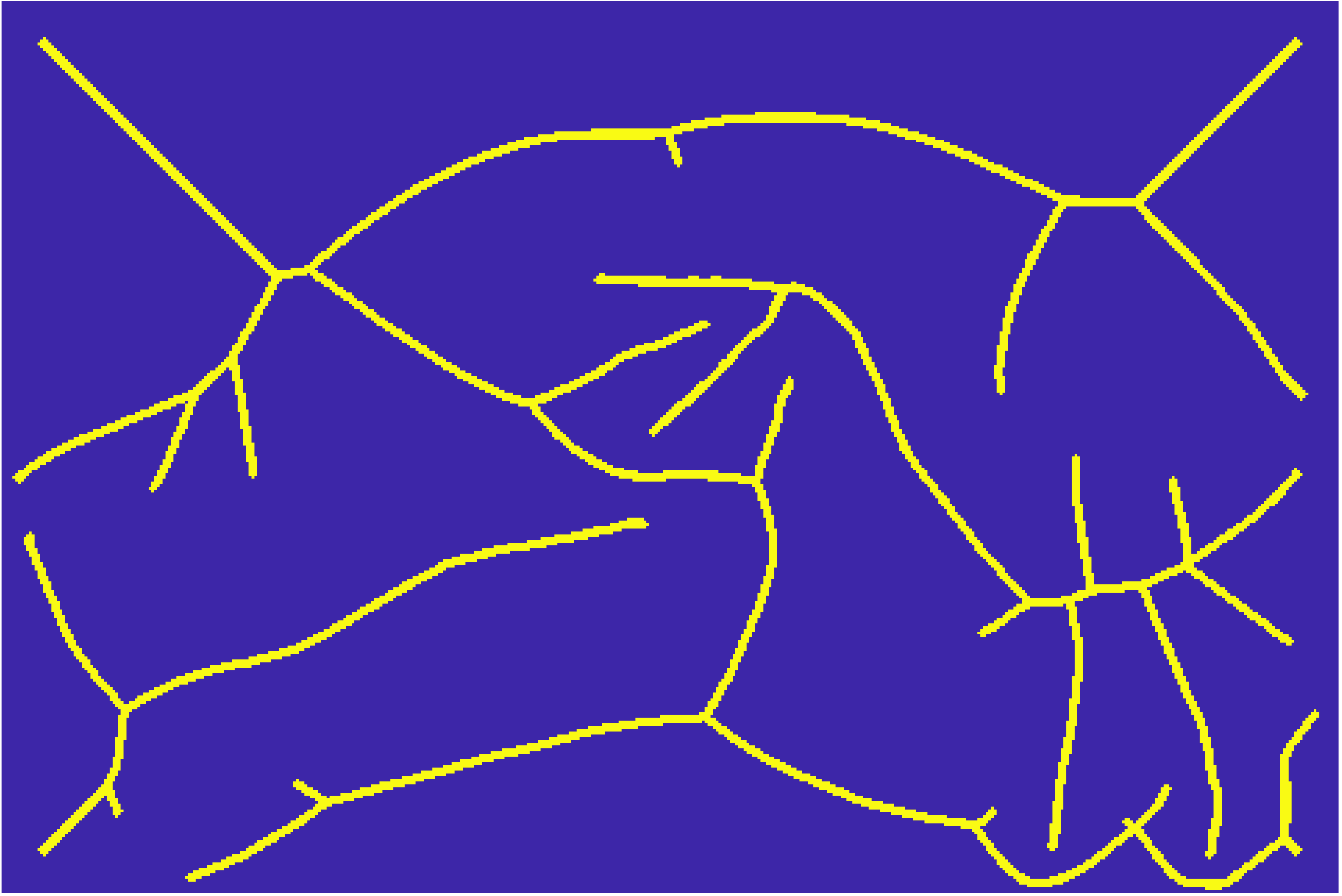}
    \includegraphics[width=\wfactor\linewidth]{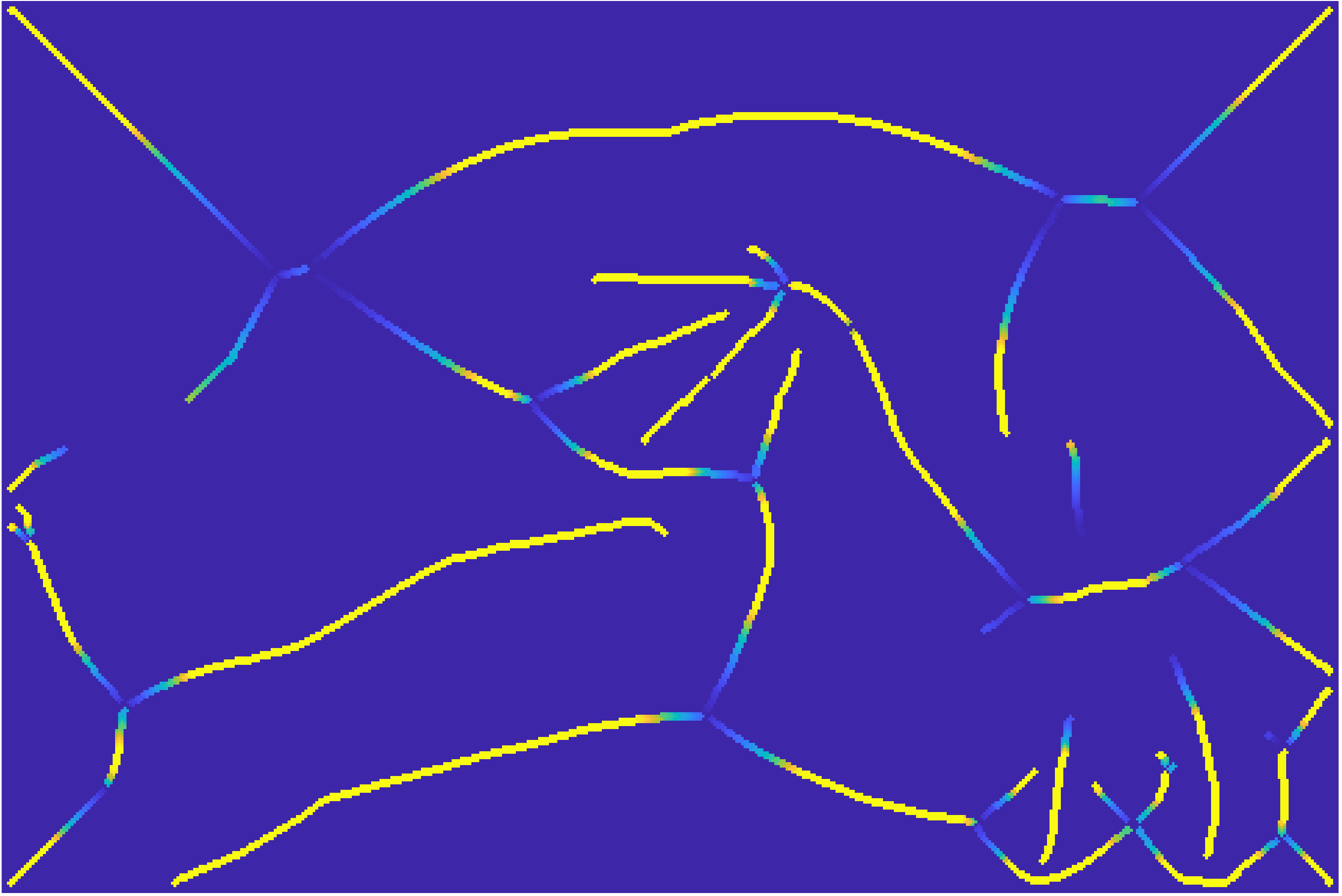}
    \caption{Segmentation (left), binary GT skeletons (middle), and their weighted 
        version based on uniqueness of medial disk area (right)~\cite{DBLP:journals/fiict/RezanejadS15}.
        The most salient skeleton parts are retained (yellow), 
        while skeletal points with low boundary support have low weights (blue).
  }
    \label{fig:weighted_groundtruth}
\end{figure}
Traditionally, the evaluation of skeleton extraction methods has followed the protocol originally introduced for the task of boundary detection
on the BSDS500 benchmark~\cite{funk2017symm}\cite{martin2001database,martin2004learning}.
According to that protocol, the extracted (boundary/skeleton) map is binarized, and then matched to each one of the 
available annotations for a given image, using a bipartite graph matching routine that allows for small localization errors.
To compute precision (P), a detected point can be matched to \emph{any} of its ground truth (GT) counterparts, while, for 
perfect recall (R), \emph{all} ground truth points must be matched with a point in the output.

We argue that this benchmarking approach can be misleading for the task of skeleton detection.
To better understand why, see~\reffig{fig:boundary_vs_skeleton_gt}.
The boundary annotations for the same scene form a natural hierarchy:
fine-grained interpretations of a scene \emph{complement} the coarser ones, resulting
in modest variation in the recall scores.
Skeleton annotations, on the other hand, not only change significantly when the source segmentation
changes, but actually \emph{conflict} with one another.
Even if a predicted skeleton perfectly matches one of the ground truths, it may be at complete odds 
with the rest, hurting the associated recall and F-score.

Although we employ the same evaluation scheme used in previous work for consistency, we propose the following alternative:
for each image, we consider each annotation individually and report scores for the one with the maximum F-score.
This is a much more reasonable expectation -- we require the output to match \emph{at least one} of the acceptable
scene interpretations, rather than all of them jointly.

Another observation we make is that large portions of the medial axis may have little to do with boundary 
reconstruction, but are due to the ligature or the ``glue'' that holds parts of the object 
together~\cite{blum1973biological}.
Curiously, all studies benchmarked on BMAX500 or SK-LARGE, have ignored this fact. 
With this in mind, we use a uniqueness of medial disk area-based ligature measure proposed by Rezanejad 
\cite{DBLP:journals/fiict/RezanejadS15}, to weight the contribution of each medial axis point 
on a scale from $[0,1]$. 
\reffig{fig:weighted_groundtruth} shows a typical example, where the 
lower weights near the branch points signal the ligature. 

\paragraph{Parameters} are optimized on the BMAX500 validation set.
We use $\alpha_c = 0.75$, $l_{max} = 10$.
$\alpha(l_F)$ is set to decrease linearly from $\alpha(1) =1$ to $\alpha(l_{max}) = 0.85$. 
We fix these values for all  experiments, including those on SK-LARGE.

We use the same values as in \cite{tsogkas2017amat} for the color cost function, 
namely $w_s = 1e-4$ and the default values for the smoothing operation \cite{xu2011image}. 
For the histogram based cost function, we use $w_s = 2e-8$. 
Finally, we set  $\rad{} \in [2,41]$ for the half-resolution images and $\rad{} \in[2,82]$ for the full resolution images.
During evaluation, any detected medial point within distance equal to $1\%$ of the image diagonal (in pixels) from the ground truth can be a true positive.

\subsection{Results} \label{sec:experiments:results}

\begin{figure*}[t!]
	\begin{tabular}{@{\hskip3pt}c@{\hskip3pt} c @{\hskip3pt}c@{\hskip3pt}}
		\fbox{\includegraphics[width=0.31\textwidth]{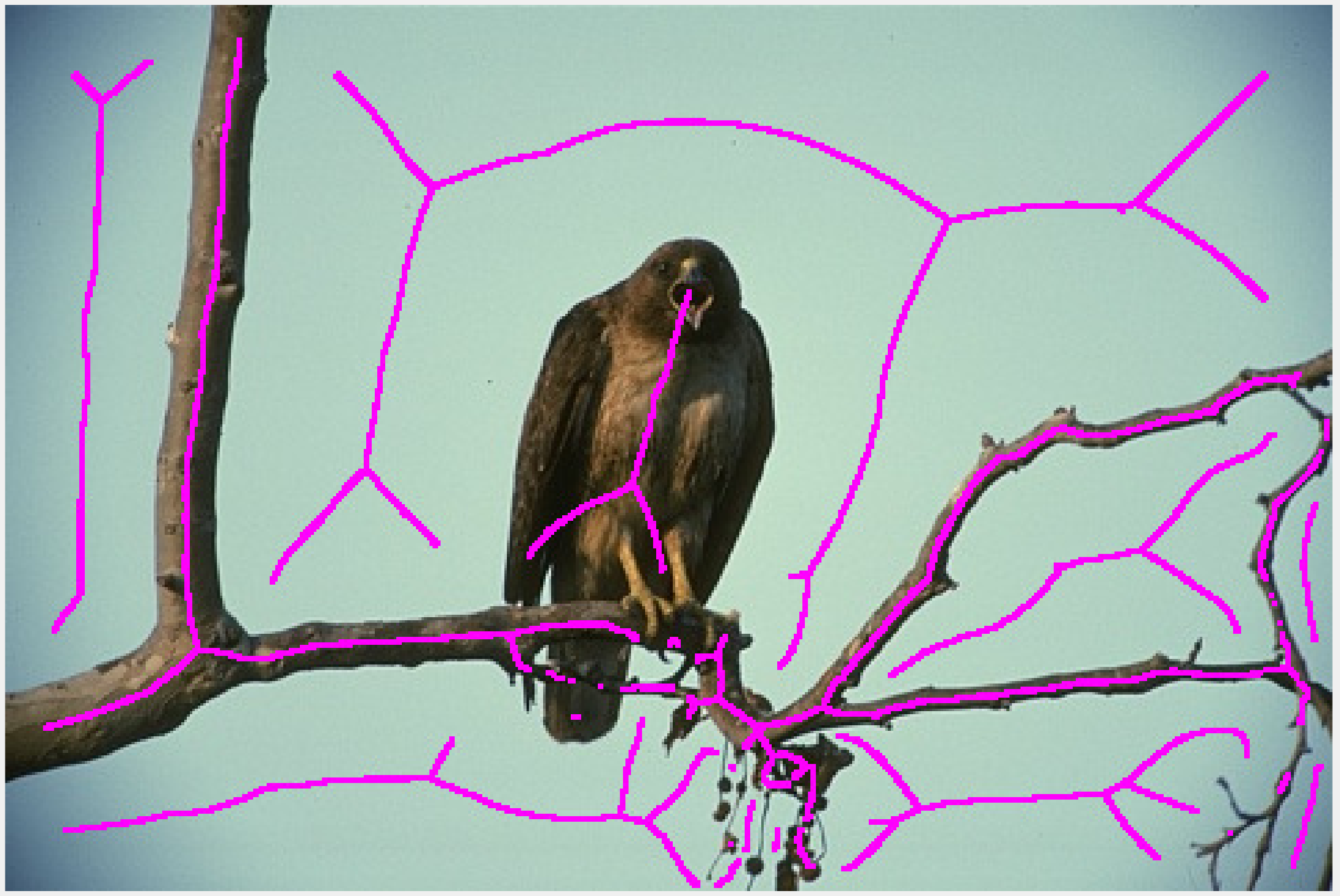}} &
		\fbox{\includegraphics[width=0.31\textwidth]{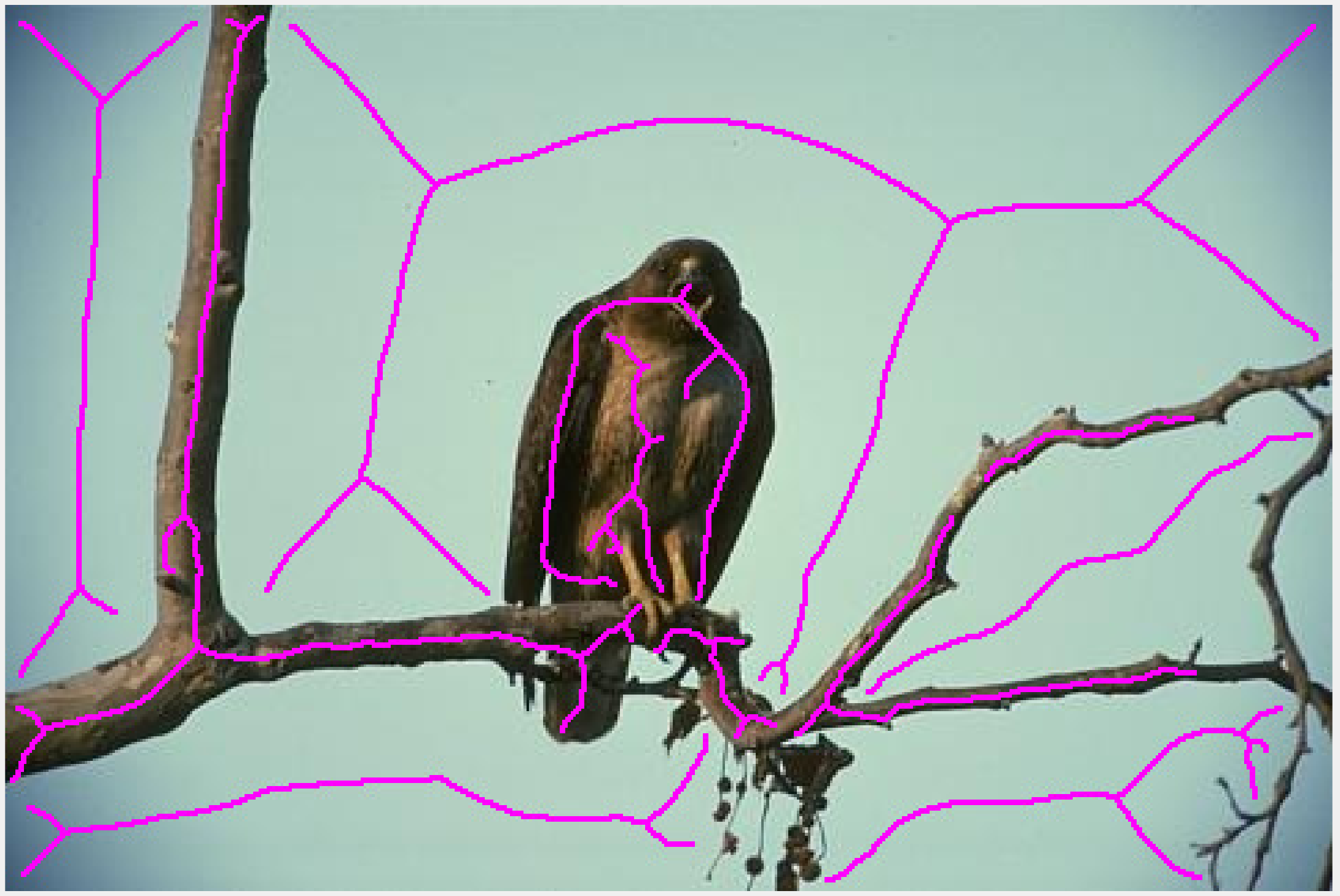}} &
		\fbox{\includegraphics[width=0.31\textwidth]{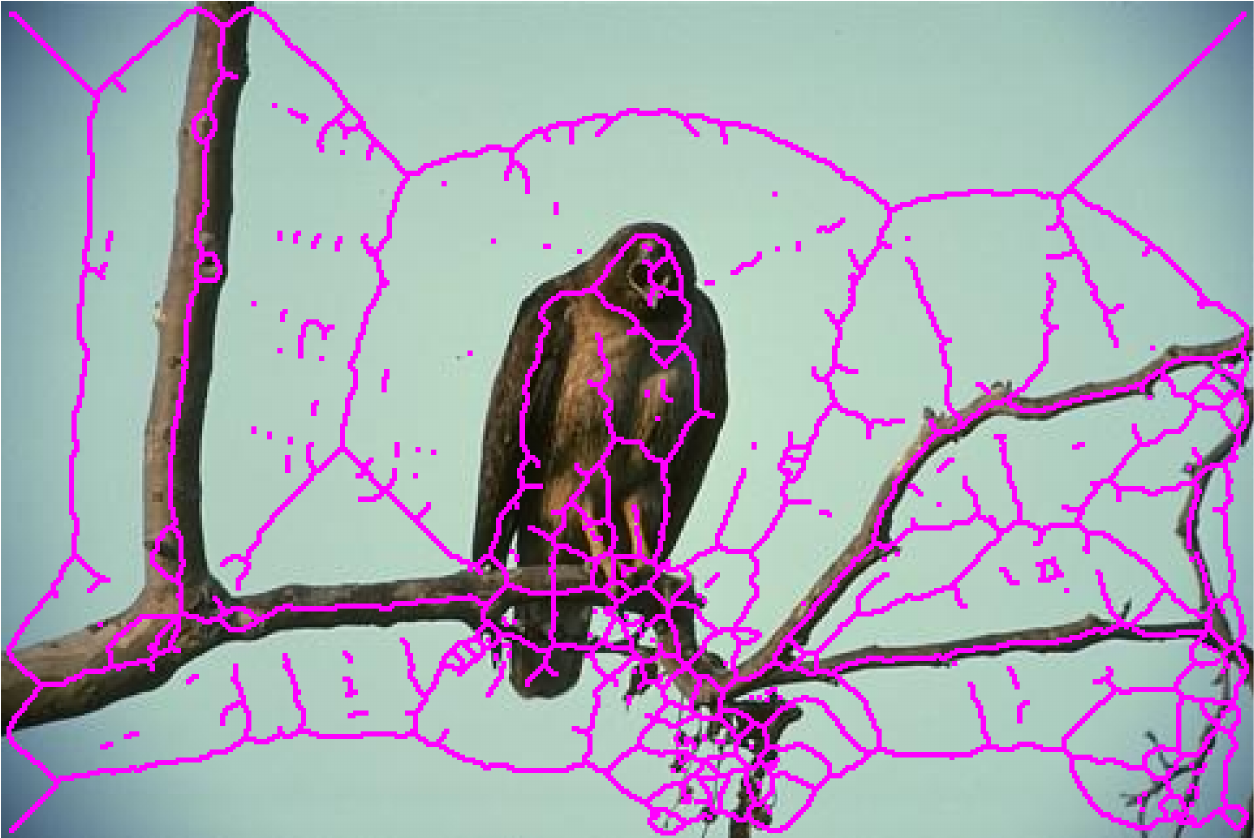}}\\
		\fbox{\includegraphics[width=0.31\textwidth]{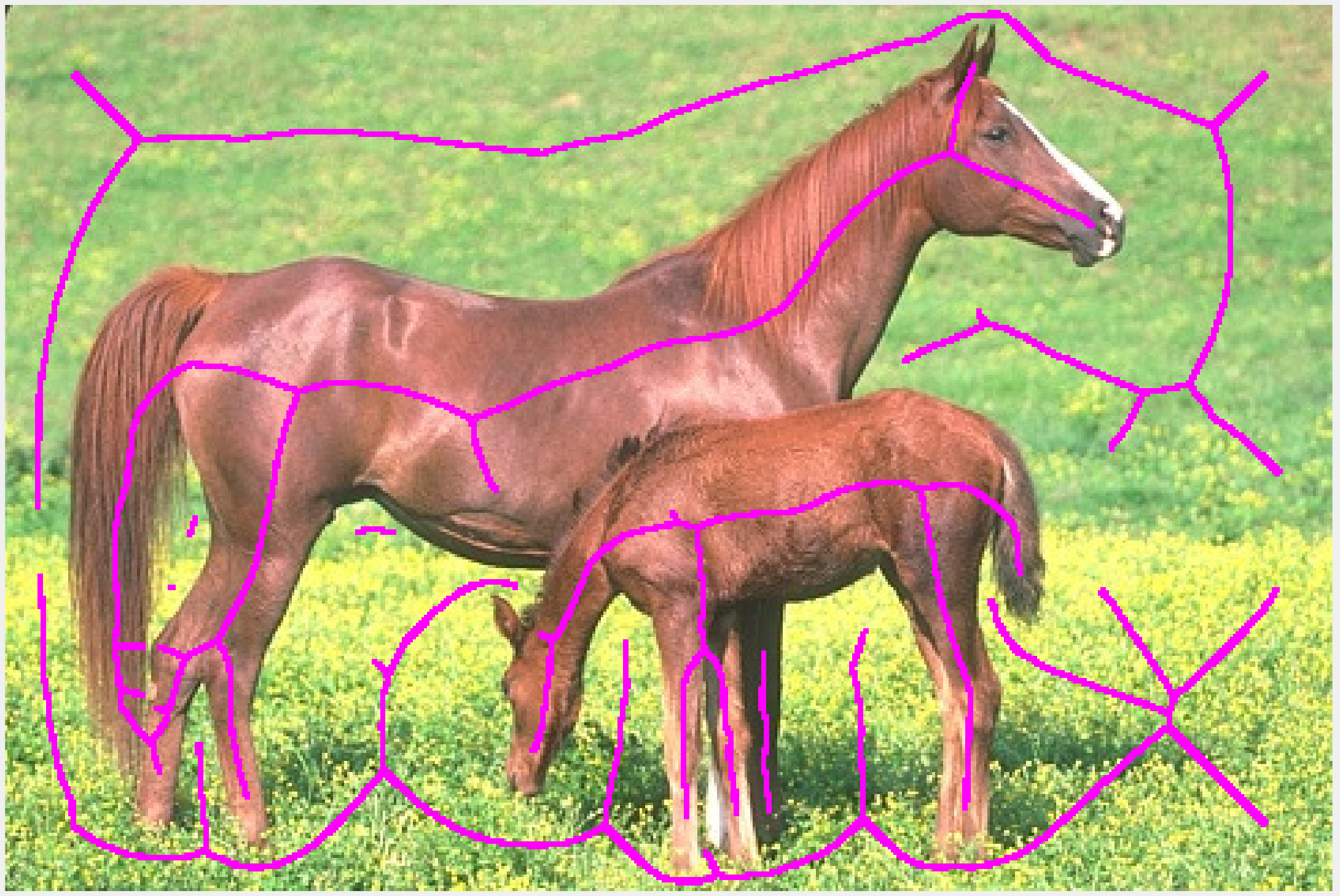}} &
		\fbox{\includegraphics[width=0.31\textwidth]{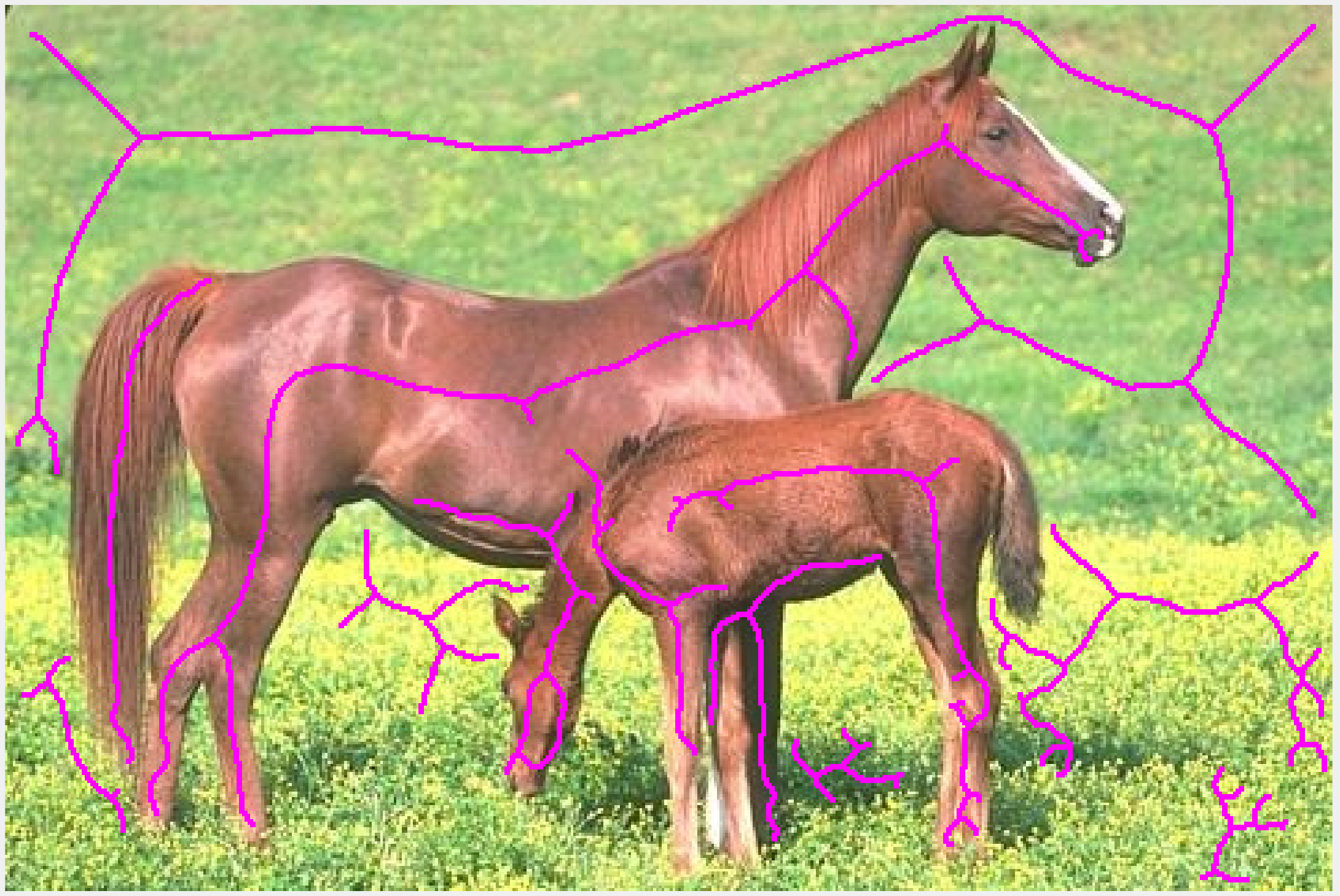}} &
		\fbox{\includegraphics[width=0.31\textwidth]{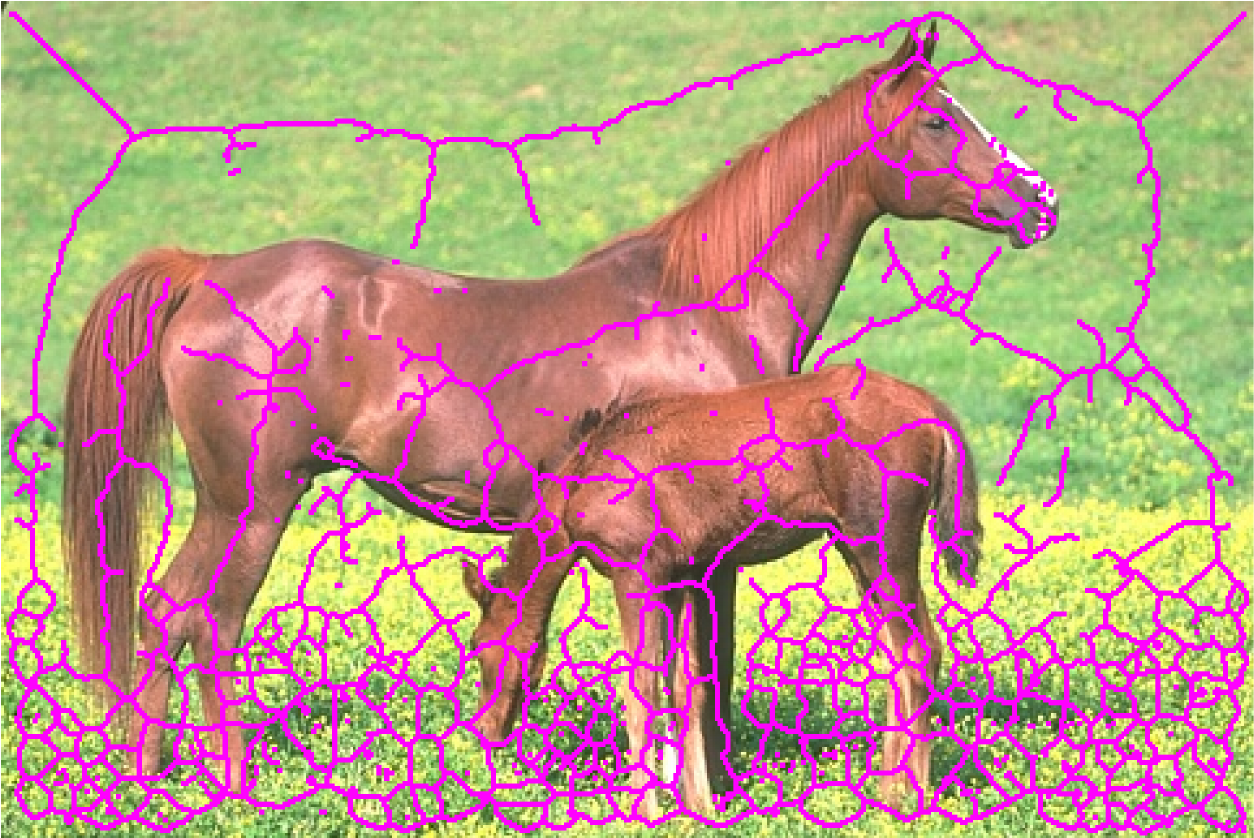}}\\
	\end{tabular}
	\caption{Qualitative results. \textbf{Left to right:} Ground truth (single annotation), 
	ASG (this work), AMAT~\cite{tsogkas2017amat} (after post-processing). 
	Our method produces sparser, cleaner, and more accurate medial axes, without any post-processing.}
	\label{fig:qualitative}
\end{figure*}
We report quantitative results for scene skeleton extraction, 
on the half- and full-resolution BMAX500 dataset, in~\reftab{tab:bmaxhalfthin}.
We compare the AMAT~\cite{tsogkas2017amat} with post-processing (i.e., grouping and thinning) and the ASG, using the two cost functions 
described in Section~\ref{sec:method:implementation}. 
We include results for both the standard and our proposed evaluation protocol, as well as 
the gains due to our ligature weighting.

\paragraph{The cost function matters.} 
Using the histogram-based cost increases performance noticeably for both the AMAT and the ASG
($+2\%$ and $+10\%$ F-score respectively).
This result confirms our hypothesis that a powerful cost function that is robust to texture and other 
local appearance variations is crucial in order to obtain good quality medial axes. 

\paragraph{Performance analysis.}
We focus on the results for the intensity-histogram cost function. 
The standard evaluation protocol rewards the AMAT's dense yet imprecise output:
predicted points have multiple ``shots'' at matching with one of the multiple GT annotations, 
and, conversely, a GT point is more likely to match a detected point. 
This increases recall, making the AMAT perform on par with our method, which produces 
a much sparser ($59\%$ fewer points at full resolution), but \emph{precise} output.
Using \emph{one} of the GT per image (blue) calibrates P/R, yielding $+5.5\%$ and $+7.6\%$
F-score for half and full resolution, respectively, and aligns quantitative results to 
what we witness qualitatively in Figure~\ref{fig:qualitative}:
a clear advantage of obeying the rules of a shock grammar in skeleton detection. 
The ASG skeletons are smoother, as singularity theory 
dictates~\cite{DBLP:journals/ijcv/GiblinK03}, 
and less sensitive to boundary artefacts, while maintaining agreement with the ground truth. 
On the contrary, the AMAT skeletons, where medial hypotheses are evaluated in isolation, 
contain spurious points and invalid branching topology.

Finally, using a ligature-weighted version of BMAX increases recall for both algorithms, 
with a net advantage for the ASG, suggesting that branches 
missed by our method tend to be less important for boundary reconstruction.

\paragraph{ASG dramatically reduces runtime.}
Comparison of the histogram variants of AMAT and ASG in \reftab{tab:bmaxhalfthin} show a speedup of $5\times$ for the latter at half-resolution 
and $11\times$ at full-resolution.
Our approach not only is faster by an order of magnitude, it also scales much better with the input image size and the 
number of scales considered.
A detailed breakdown of the algorithm  is shown in~\reftab{tab:runtime}.

\begin{table}
    \resizebox{\linewidth}{!}{
    \centering
    \begin{tabular}{c|c|c||c|c}
        Resolution  & \multicolumn{2}{c||}{$161\times241$} & \multicolumn{2}{c}{$321\times481$} \\
        \hline
        Proposal Generation  & 3.63s & 36.0\%   & 63.51s &  64.7\% \\
        Seed Growth  & 4.60s & 45.6\% & 18.28s & 18.6\%\\
        End Point Growth  & 1.85s & 18.3\% & 15.71s & 15.9\%\\
         Other  & 0.01s &  0.1\% & 0.73s & 0.8\% \\\hline
        Total & 10.09s & 100\% & 98.23s & 100\% \\
    \end{tabular}
    }
    \caption{Runtime breakdown of the ASG. Timings are averages over the 200 images in 
        the BMAX500 test set.
        \emph{Other} includes the seed extraction, selection, and pruning steps.
    }
    \label{tab:runtime}
\end{table}

\paragraph{Comparison with supervised methods.}
In \reftab{tab:sklarge} we compare to supervised learning methods.
SK-LARGE contains annotations only for the foreground object skeletons, 
so we ignore medial axes outside the object during evaluation.
Both the AMAT and ASG produce lower F-scores than 
Hi-Fi~\cite{zhao2018hi} and DeepFlux~\cite{wang2018deepflux},
but this is expected because they are not solving the same problem:
the former rely solely on bottom-up features to extract medial axes of \emph{homogeneous image regions}, 
whereas the latter incorporate high-level, object-specific information to 
detect \emph{semantic object skeletons}. 
Taking these numbers at face value also ignores many ``hidden costs'' of supervised learning: 
1) training deep CNNs for skeleton extraction requires GPUs and segmentations, 
which are costly and time consuming to collect; 
2) these models do not generalize on other datasets: \cite{tsogkas2017amat} showed that 
FSDS~\cite{shen2016object}, trained on SK-LARGE, fails to generalize on BMAX500; and 
3) they do not easily scale to new classes or granularities; 
e.g., if a new class is added to the dataset, the model must be retrained.
\begin{table}[!t]
    \setlength\tabcolsep{3pt}
    \resizebox{\linewidth}{!}{
        \begin{tabular}{c| c c||cc}
            & AMAT~\cite{tsogkas2017amat} & ASG & Hi-Fi~\cite{zhao2018hi} & DeepFlux~\cite{wang2018deepflux} \\
            \hline
            F1 & .509 & \bf{.511} & .724 & \bf{.732} \\     
            t (s) & 511.9 & \bf{63.2} & 0.030 (GPU) & \bf{0.019} (GPU) \\
        \end{tabular}
    }
    \caption{Results on SK-LARGE~\cite{shen2017deepskeleton}. 
        Runtimes are averages over the SK-LARGE test images.
    }
    \label{tab:sklarge}
\end{table}

%% file: discussion.tex
\section{Discussion}\label{sec:discussion}
Our new approach for the efficient extraction of medial axes from cluttered natural scenes
uses elements from the shock graph theory of shape.
In particular, we have generalized the concept of shocks to the RGB domain by considering region-based cost functions, 
and have devised an algorithm that leverages the rules of the shock graph grammar to guide the search for medial points.
Our approach has several merits: 
1) it is fully \emph{unsupervised} and thus can generalize to new datasets without any training; 
2) it outperforms the state-of-the-art in unsupervised approaches and
is an order of magnitude faster and much more efficient in the number of skeletal pixels generated; and 
3) it requires no post-processing, such as thinning or grouping of medial points.
In our experiments, we have also raised a concern regarding the way scene skeleton detection frameworks are typically evaluated in our community. 
To address this, we have proposed an alternative, ligature-based weighted evaluation scheme, that
takes into account the relative importance of each medial point for boundary reconstruction, 
and better reflects performance on benchmarks with multiple ground truth annotations per scene.

%% file: acknowledgements.tex
\section*{Acknowledgements}
We thank the Natural Sciences and Engineering Research Council of Canada (NSERC), 
the Fonds de recherche du Qu\'{e}bec - Nature et technologies (FRQNT), and Samsung for research funding.

%% file: supplemental.tex
\appendix
\onecolumn{
\section{Appendix}
We include several additional examples from the BMAX500 dataset to highlight the benefits of using the rules of a shock grammar in unsupervised medial axis extraction from real images. 
Here we show the ground truth medial axis (left), our new ASG result (middle) and the AMAT result after post-processing (right). 
For both the ASG and the AMAT we use our histogram of intensity cost function. 
The ASG generally gives smoother, more complete medial axes that are also more consistent with the underlying region textures. }
\begin{figure*}[!hbt]
	\begin{tabular}{@{\hskip3pt}c@{\hskip3pt} c @{\hskip3pt}c@{\hskip3pt}}
		\fbox{\includegraphics[width=0.31\textwidth]{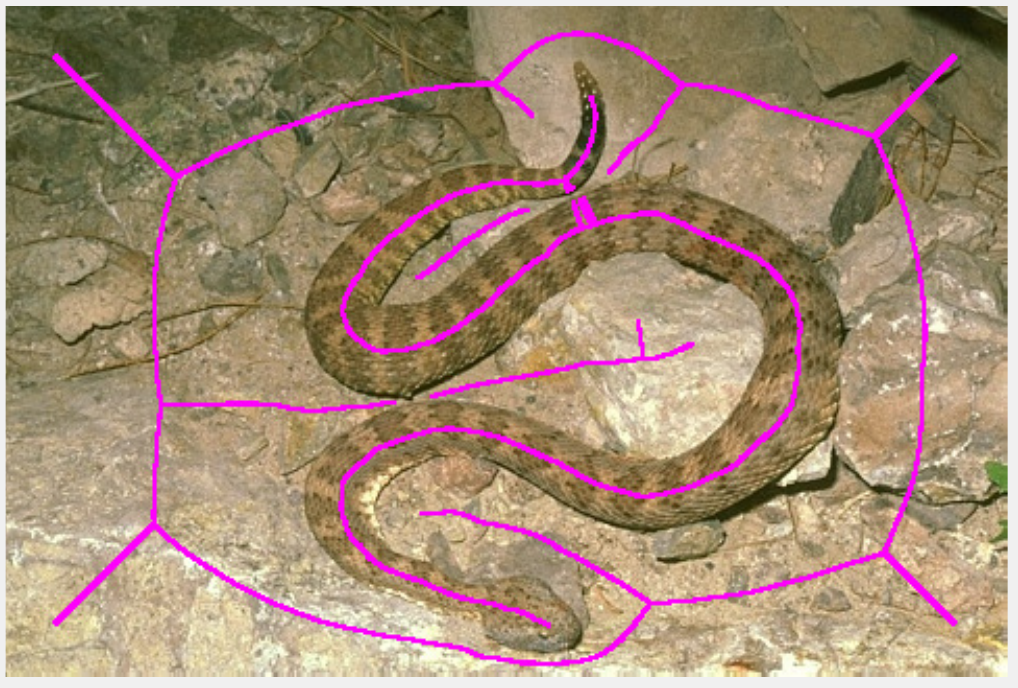}} &
		\fbox{\includegraphics[width=0.31\textwidth]{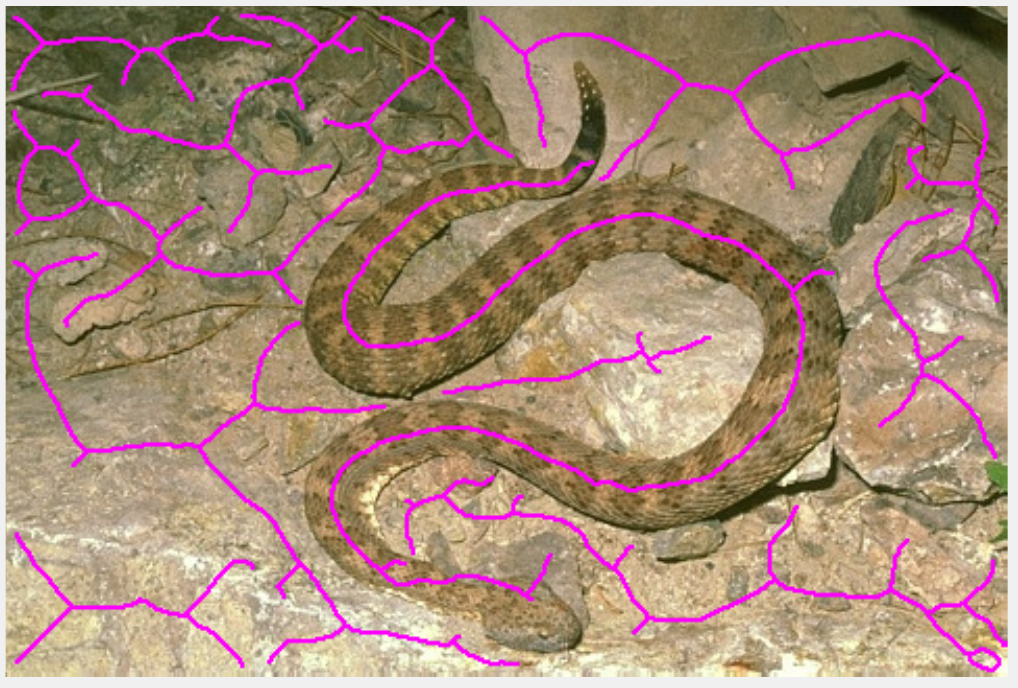}} &
		\fbox{\includegraphics[width=0.31\textwidth]{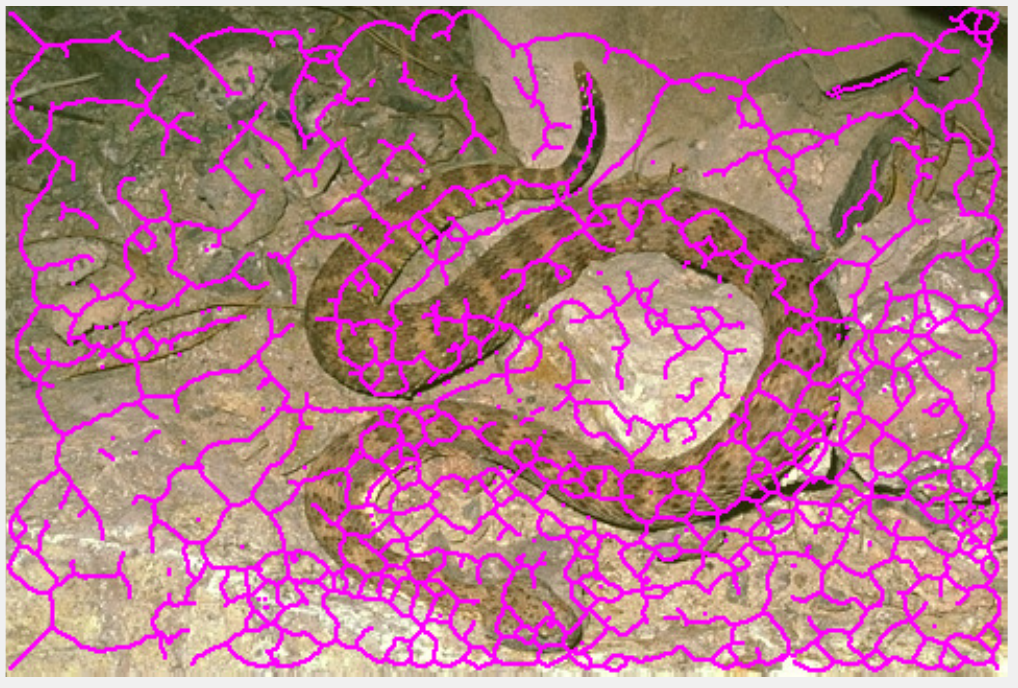}}\\
		\fbox{\includegraphics[width=0.31\textwidth]{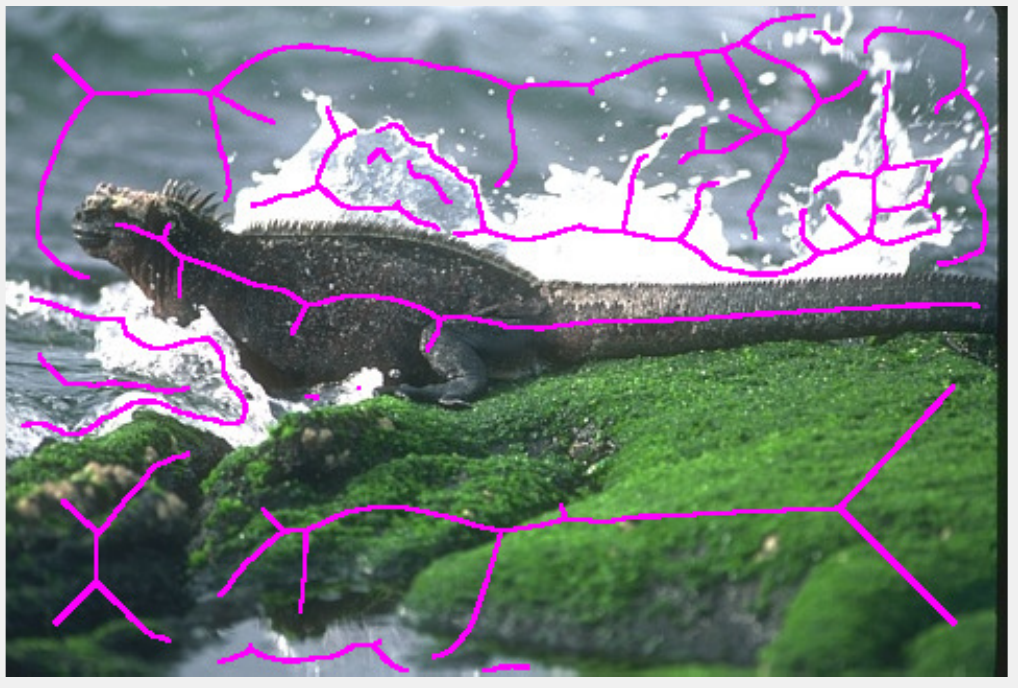}} &
		\fbox{\includegraphics[width=0.31\textwidth]{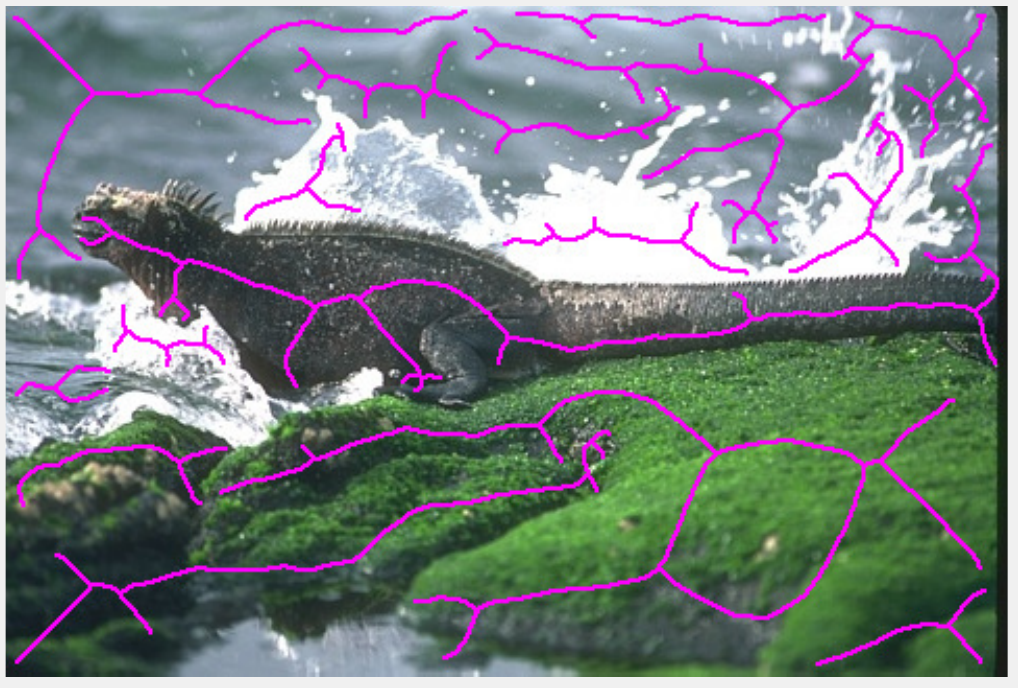}} &
		\fbox{\includegraphics[width=0.31\textwidth]{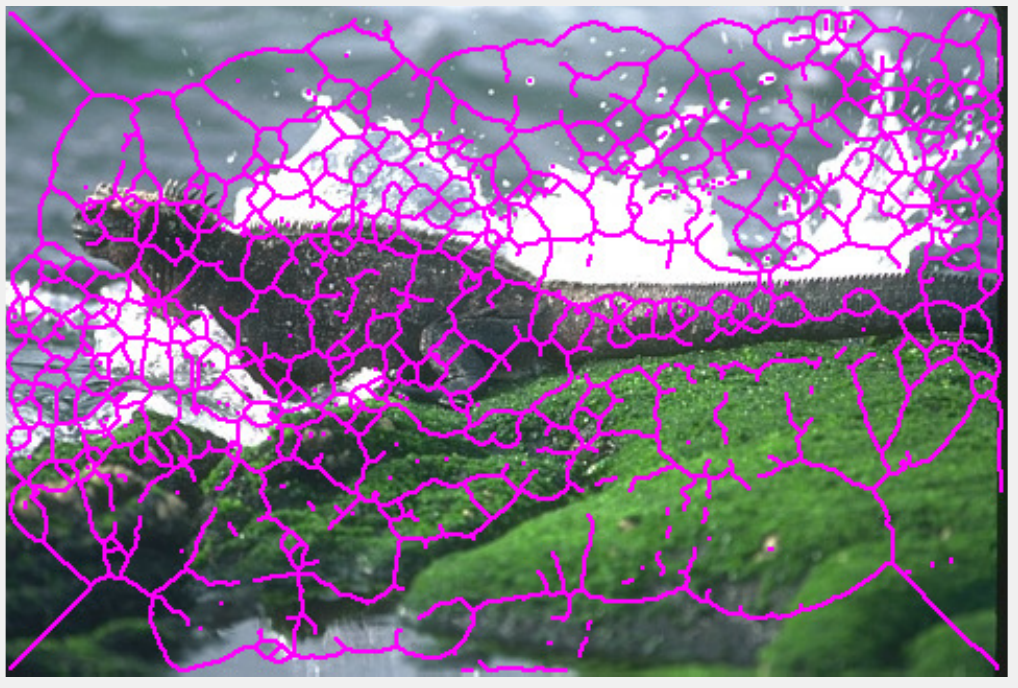}}\\
 		\fbox{\includegraphics[width=0.31\textwidth]{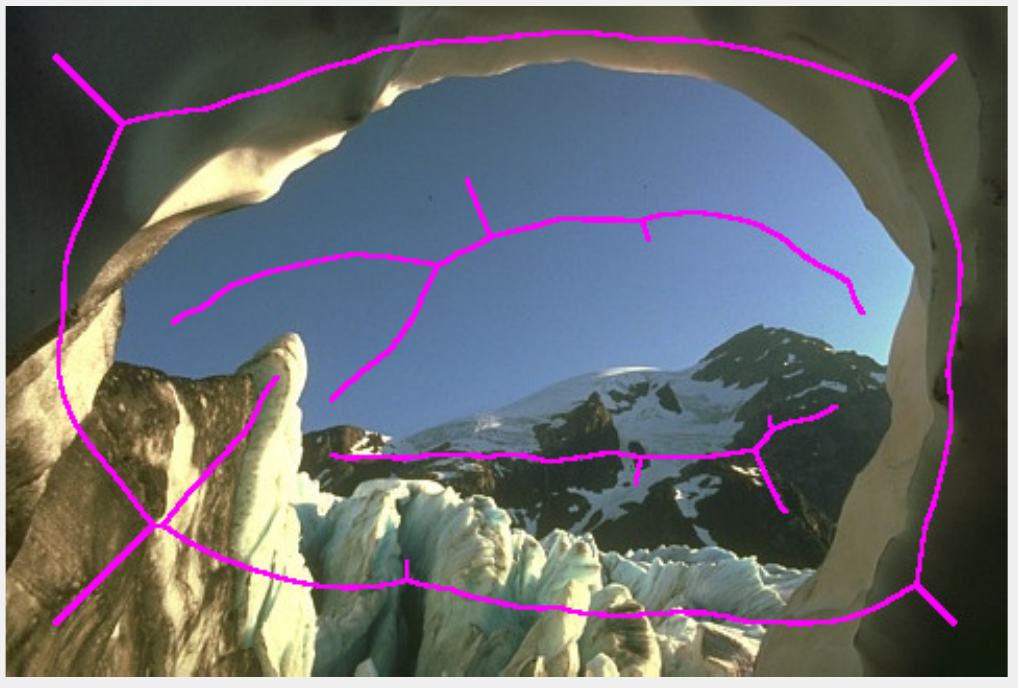}} &
 		\fbox{\includegraphics[width=0.31\textwidth]{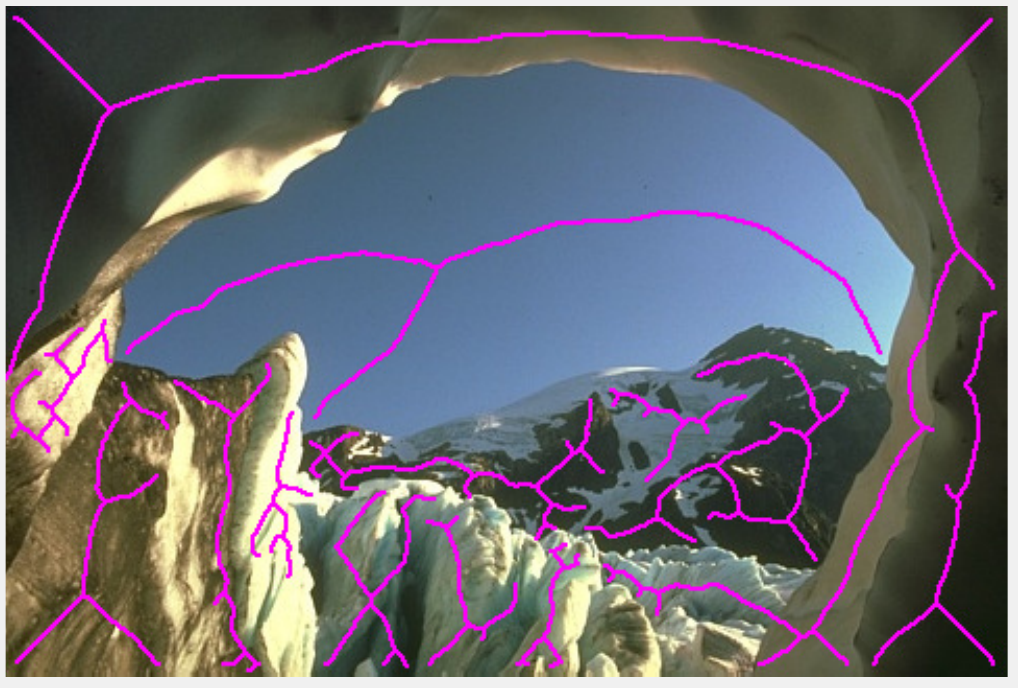}} &
 		\fbox{\includegraphics[width=0.31\textwidth]{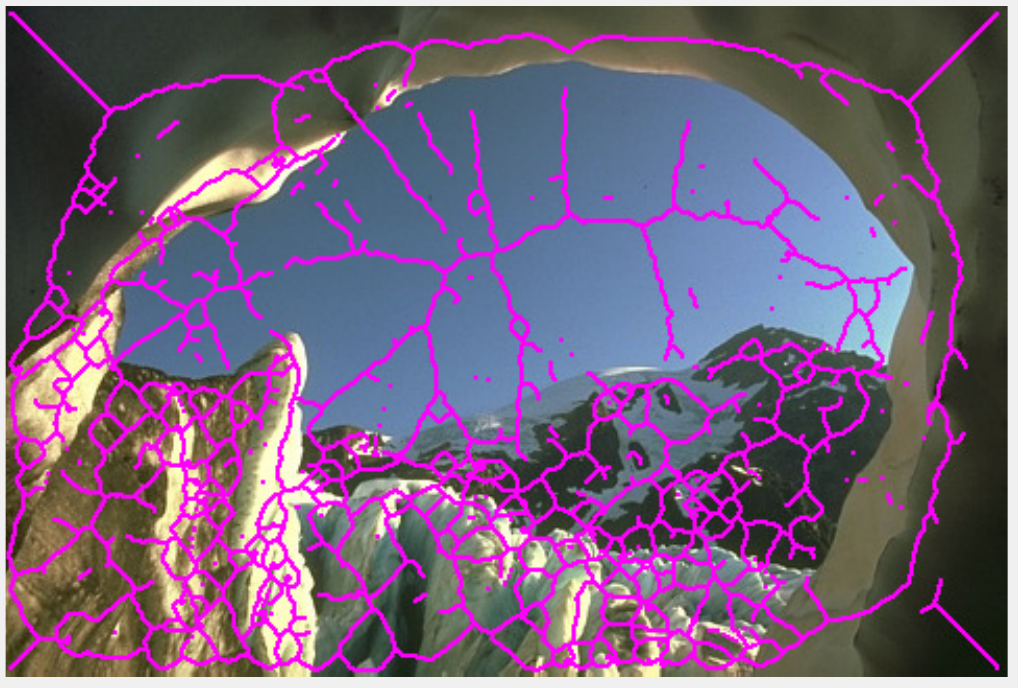}}\\
		
	\end{tabular}
	\caption{Qualitative results. \textbf{Left to right:} Ground-truth (1 annotation), ASG result, AMAT result. }
	\label{fig:qualitatives1}
\end{figure*}

\begin{figure*}[!hbt]
	\begin{tabular}{@{\hskip3pt}c@{\hskip3pt} c @{\hskip3pt}c@{\hskip3pt}}
		\fbox{\includegraphics[width=0.31\textwidth]{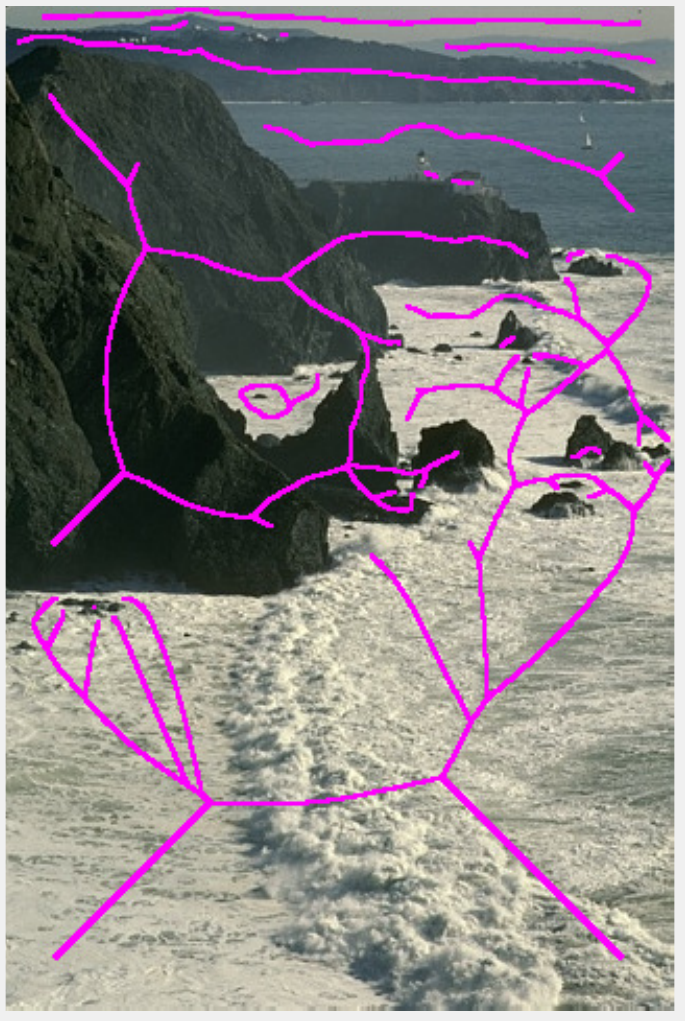}} &
		\fbox{\includegraphics[width=0.31\textwidth]{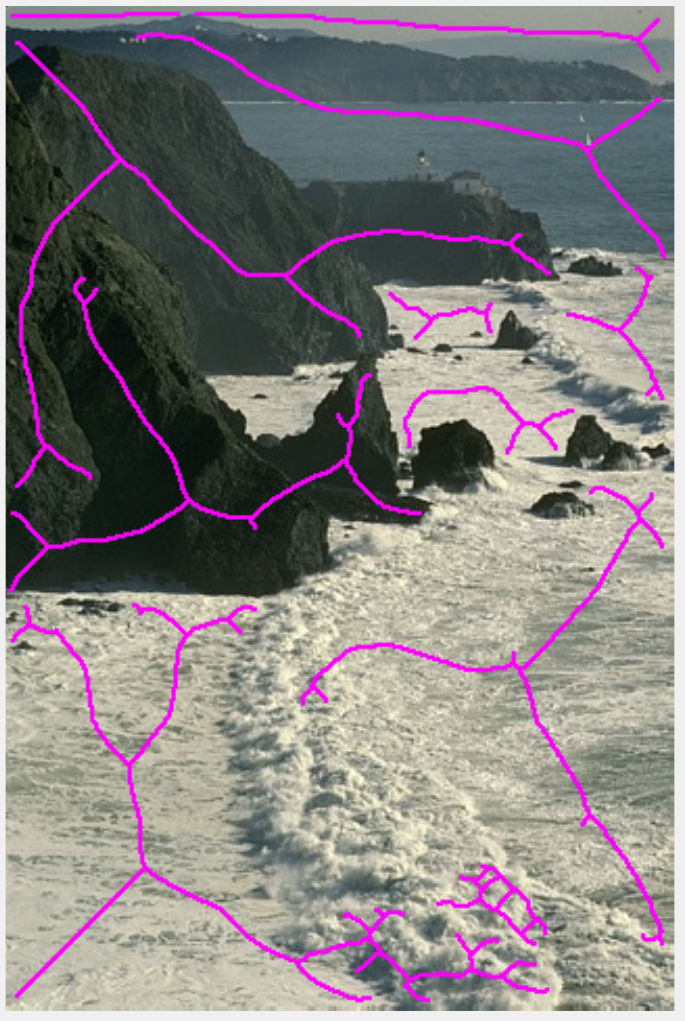}} &
		\fbox{\includegraphics[width=0.31\textwidth]{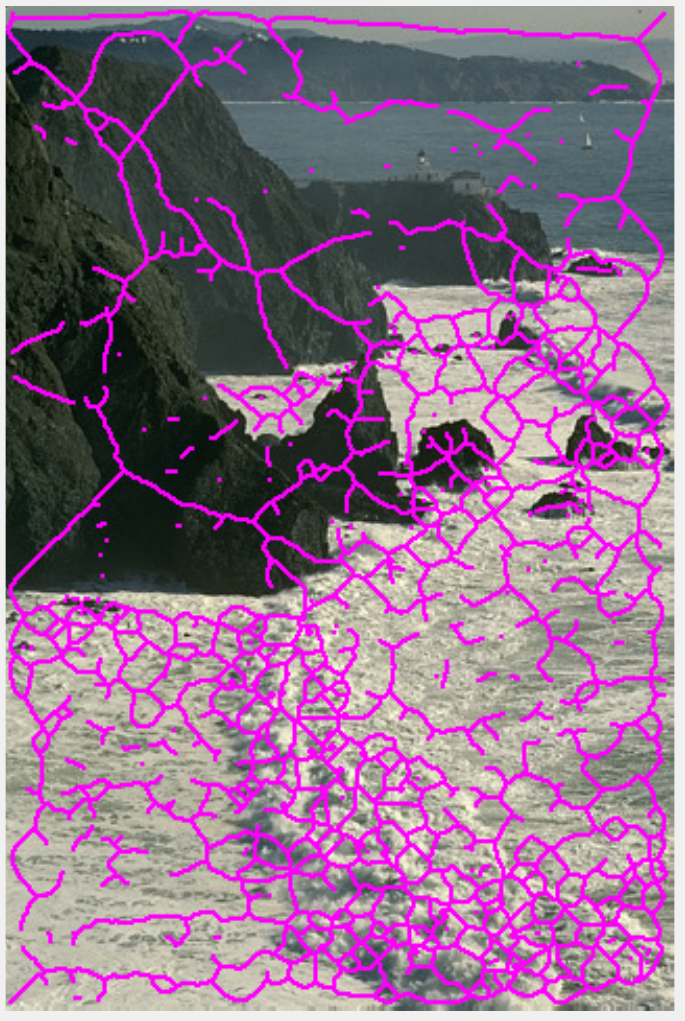}}\\
				\fbox{\includegraphics[width=0.31\textwidth]{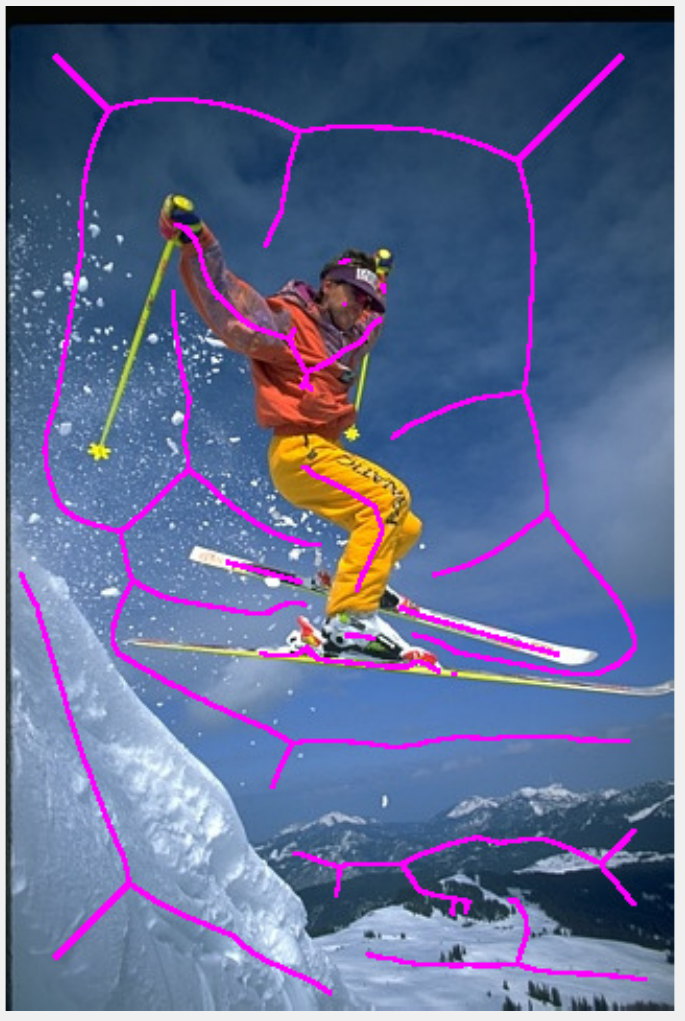}} &
		\fbox{\includegraphics[width=0.31\textwidth]{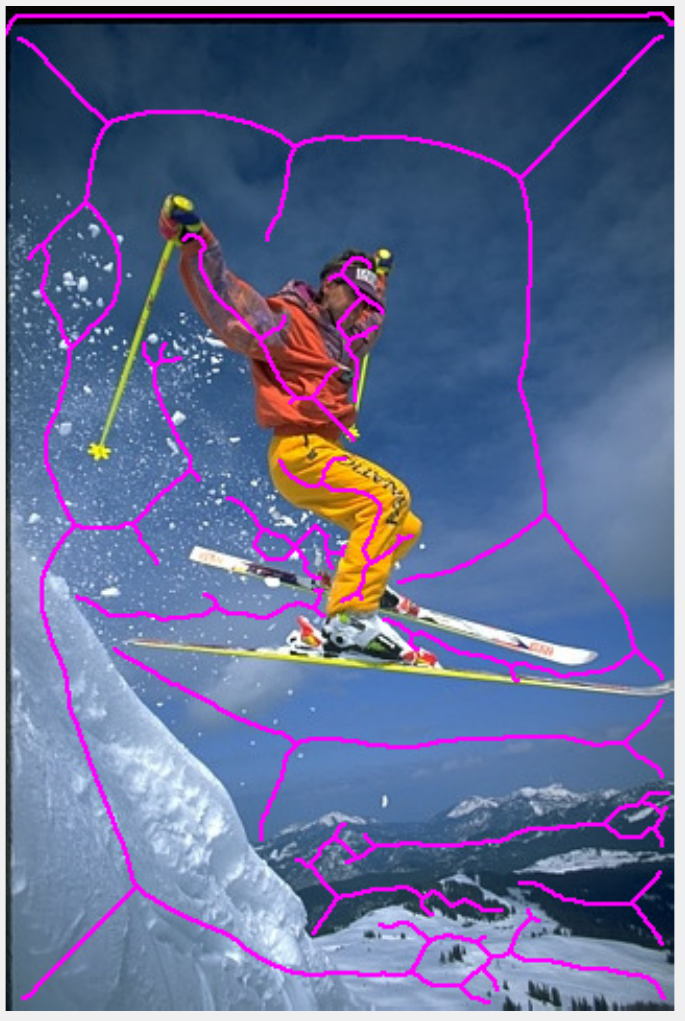}} &
		\fbox{\includegraphics[width=0.31\textwidth]{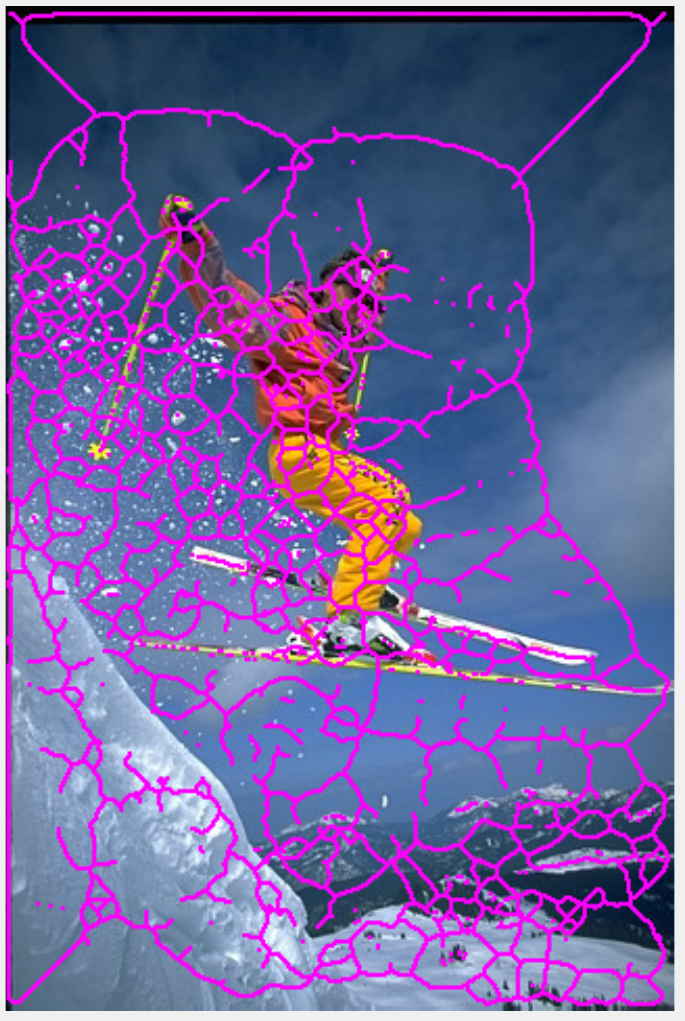}}\\
		\fbox{\includegraphics[width=0.31\textwidth]{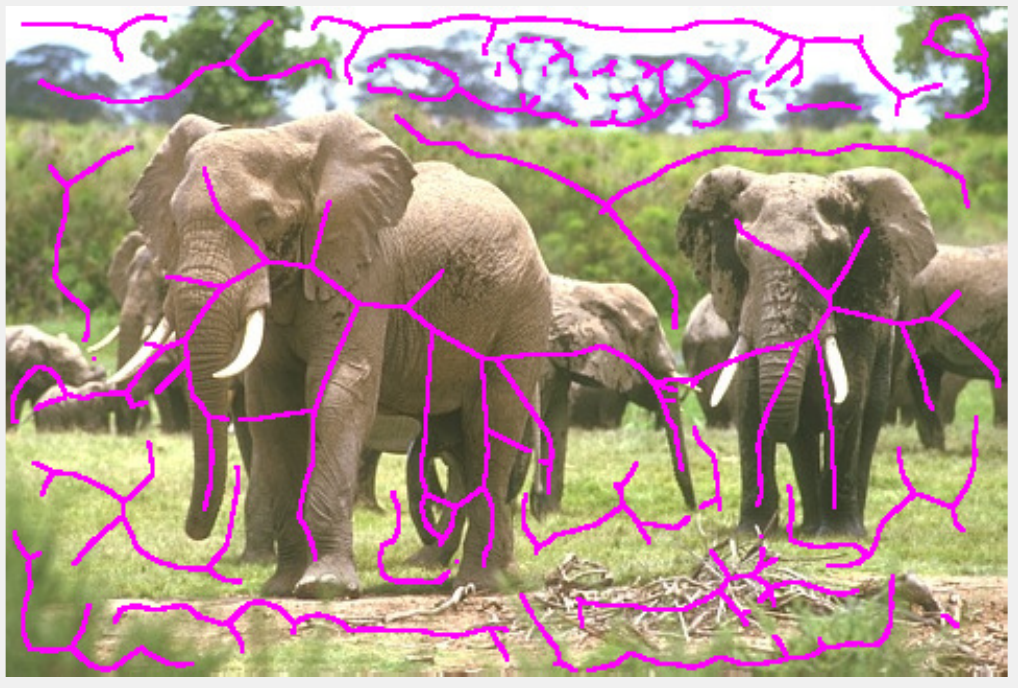}} &
		\fbox{\includegraphics[width=0.31\textwidth]{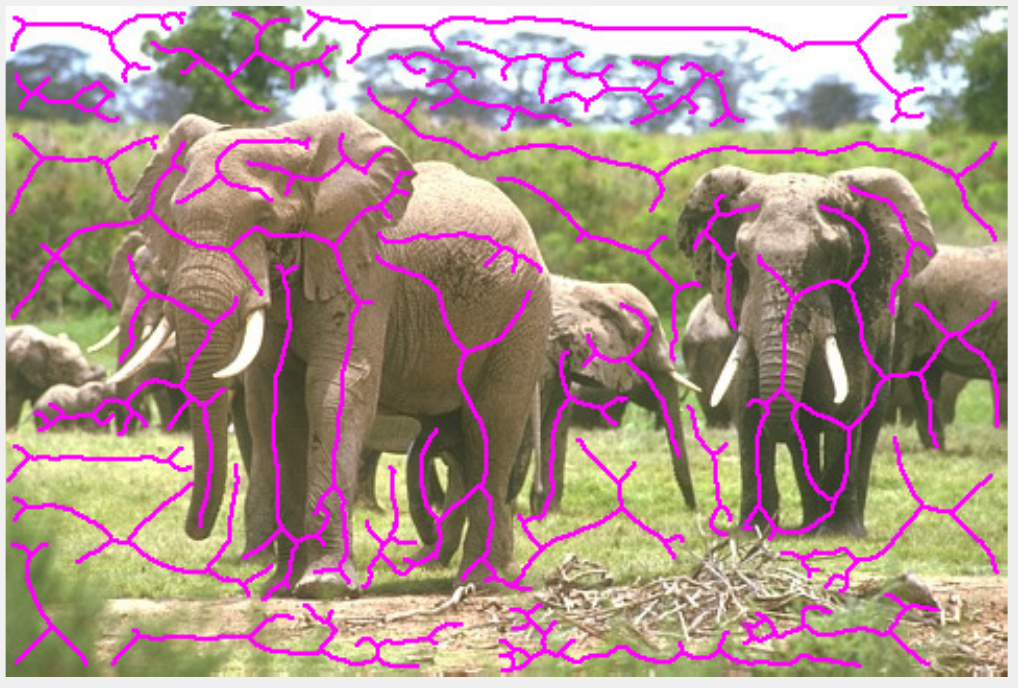}} &
		\fbox{\includegraphics[width=0.31\textwidth]{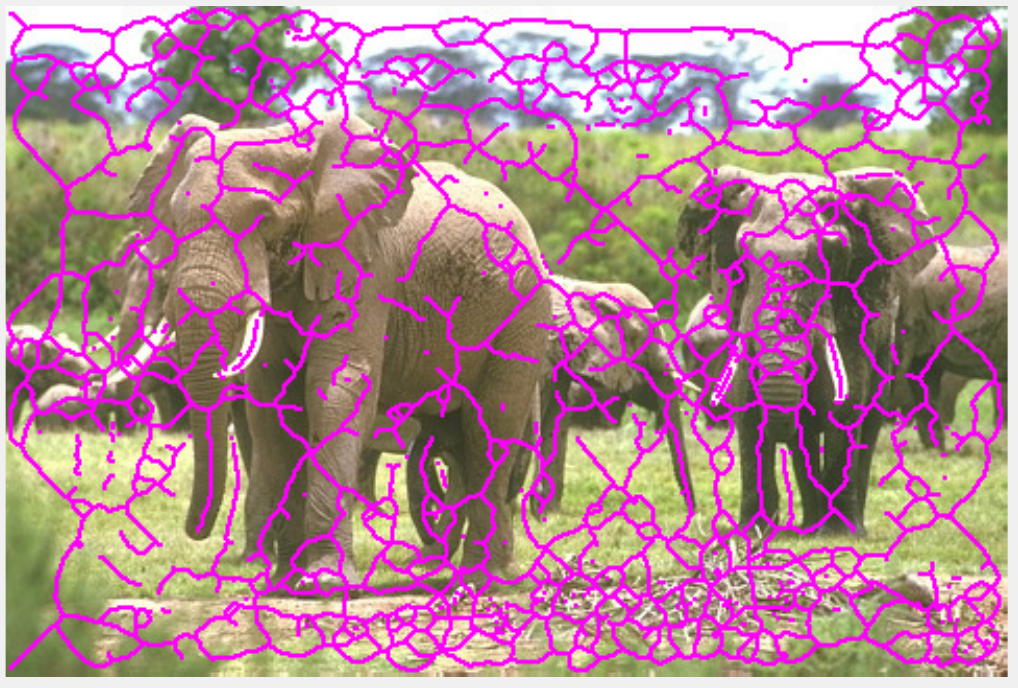}}\\
	\end{tabular}
	\caption{Qualitative results. \textbf{Left to right:} Ground-truth (1 annotation), ASG result, AMAT result. }
	\label{fig:qualitatives2}
\end{figure*}

\begin{figure*}[!hbt]
	\begin{tabular}{@{\hskip3pt}c@{\hskip3pt} c @{\hskip3pt}c@{\hskip3pt}}

		\fbox{\includegraphics[width=0.31\textwidth]{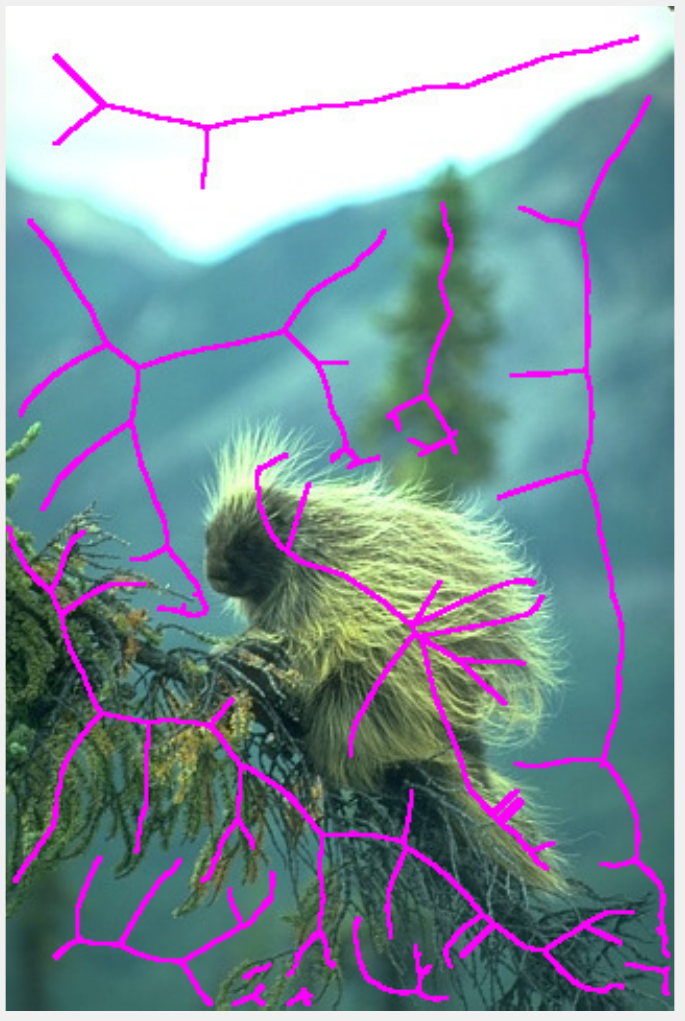}} &
		\fbox{\includegraphics[width=0.31\textwidth]{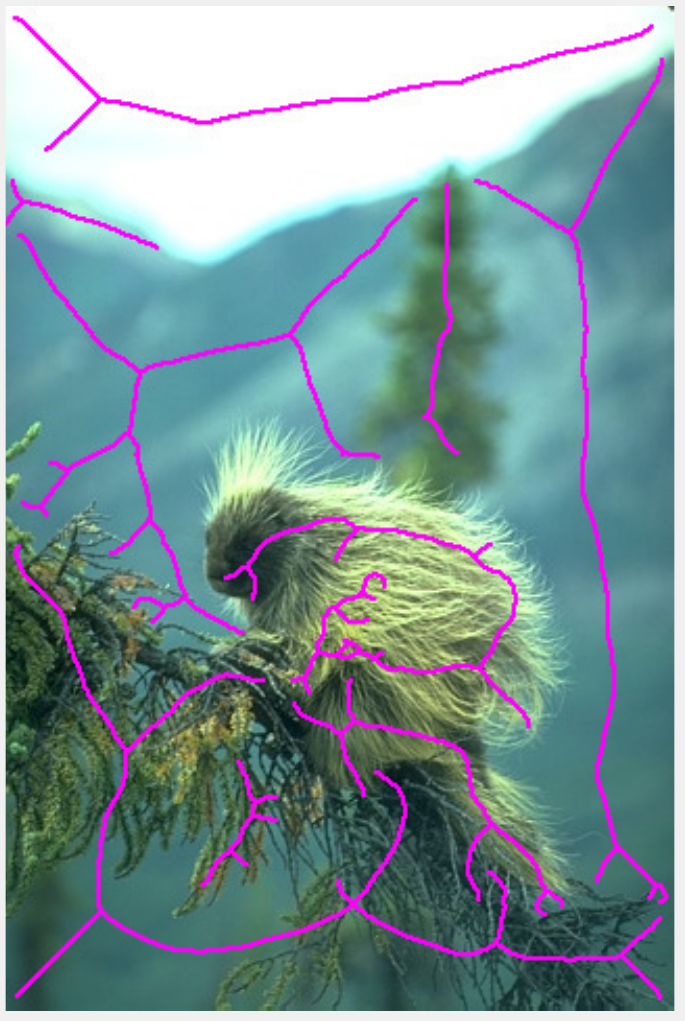}} &
		\fbox{\includegraphics[width=0.31\textwidth]{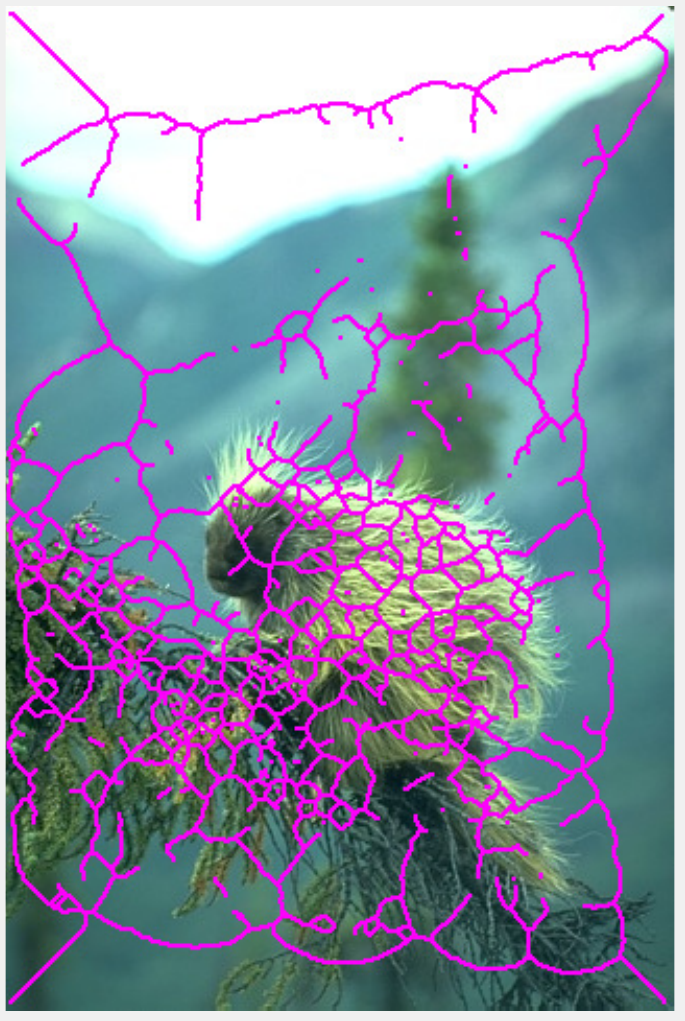}}\\
		\fbox{\includegraphics[width=0.31\textwidth]{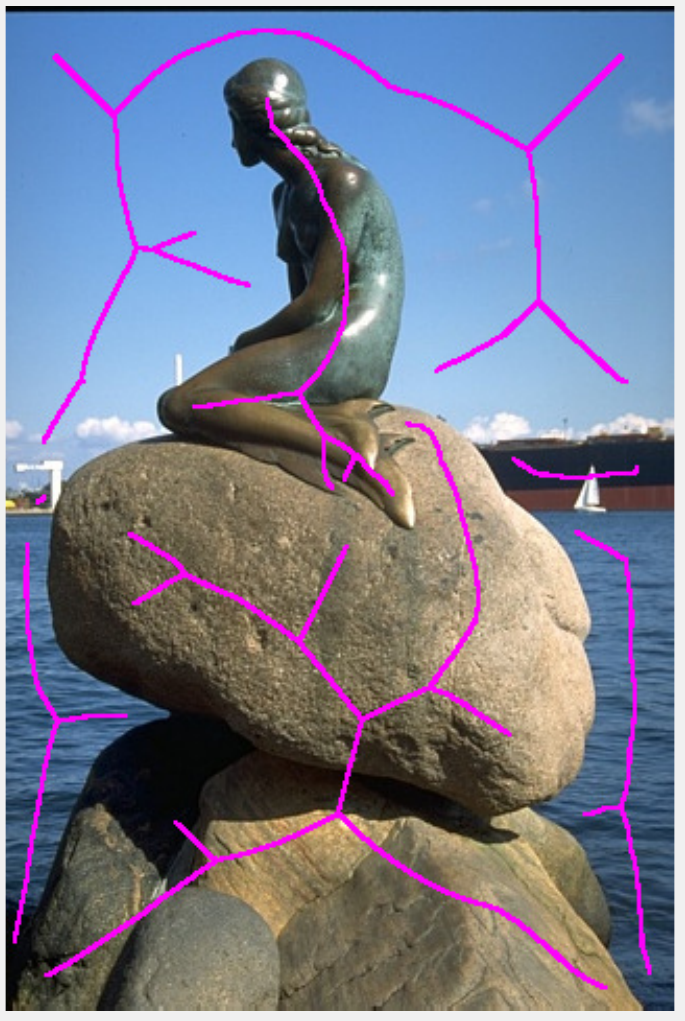}} &
		\fbox{\includegraphics[width=0.31\textwidth]{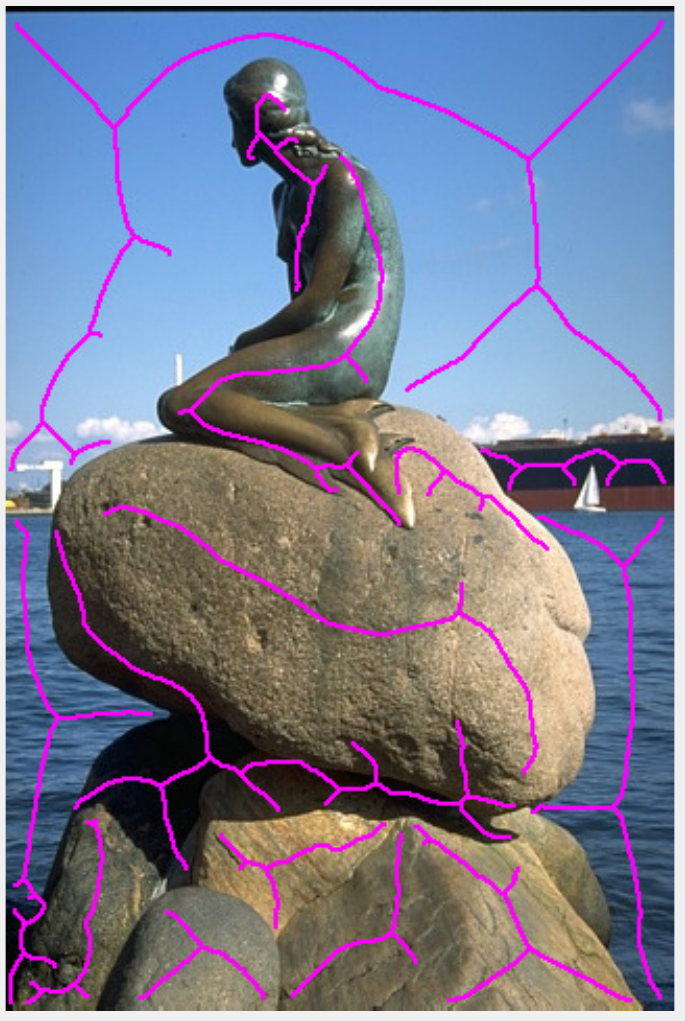}} &
		\fbox{\includegraphics[width=0.31\textwidth]{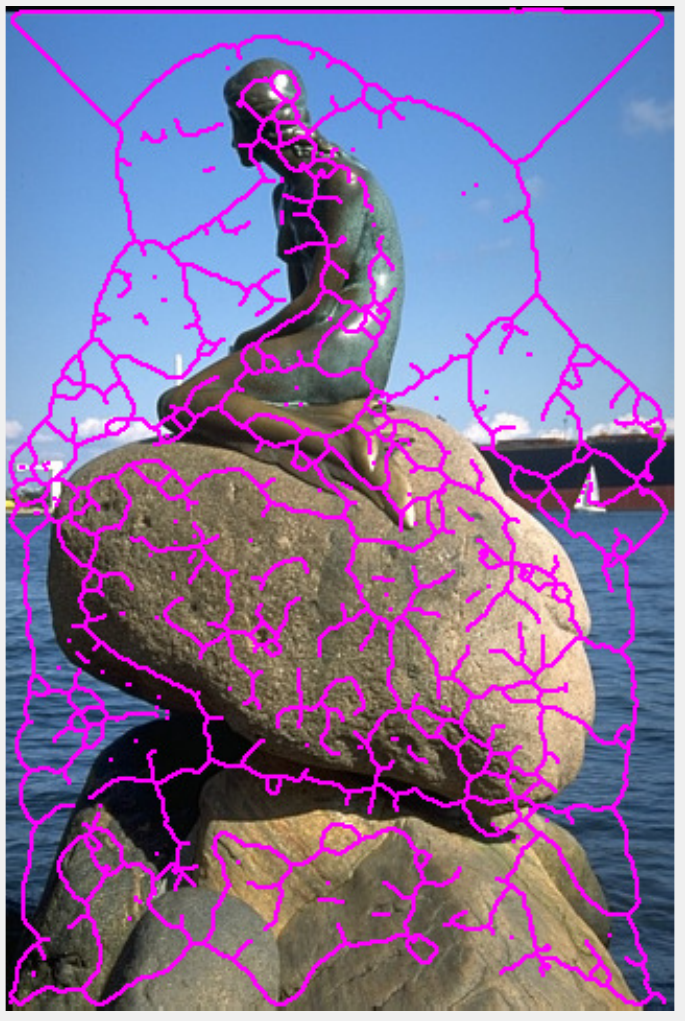}}\\
		\fbox{\includegraphics[width=0.31\textwidth]{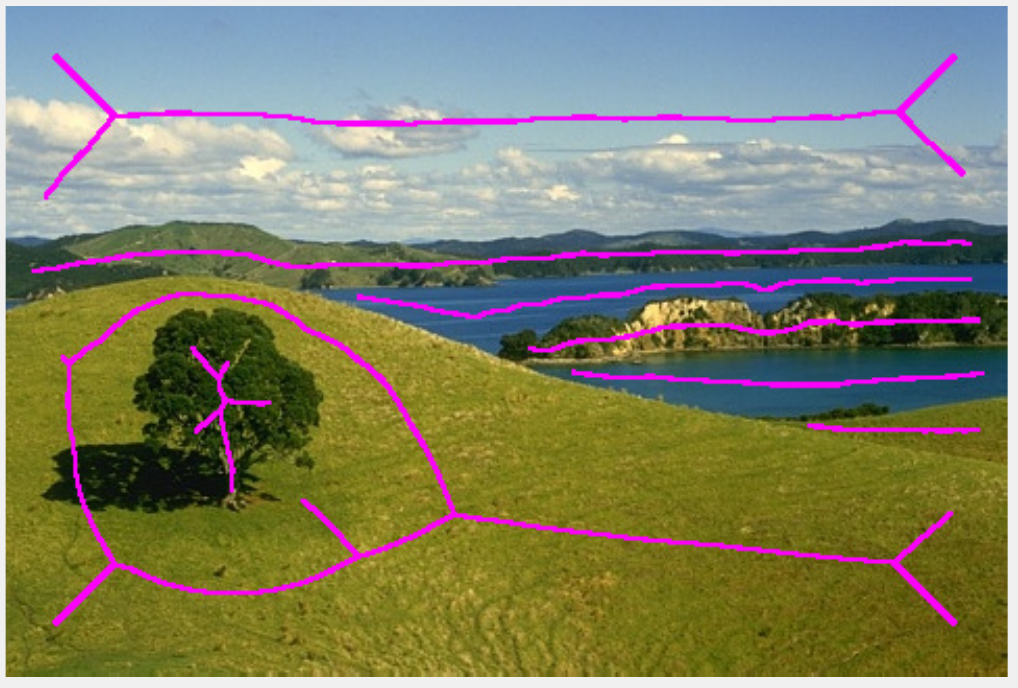}} &
		\fbox{\includegraphics[width=0.31\textwidth]{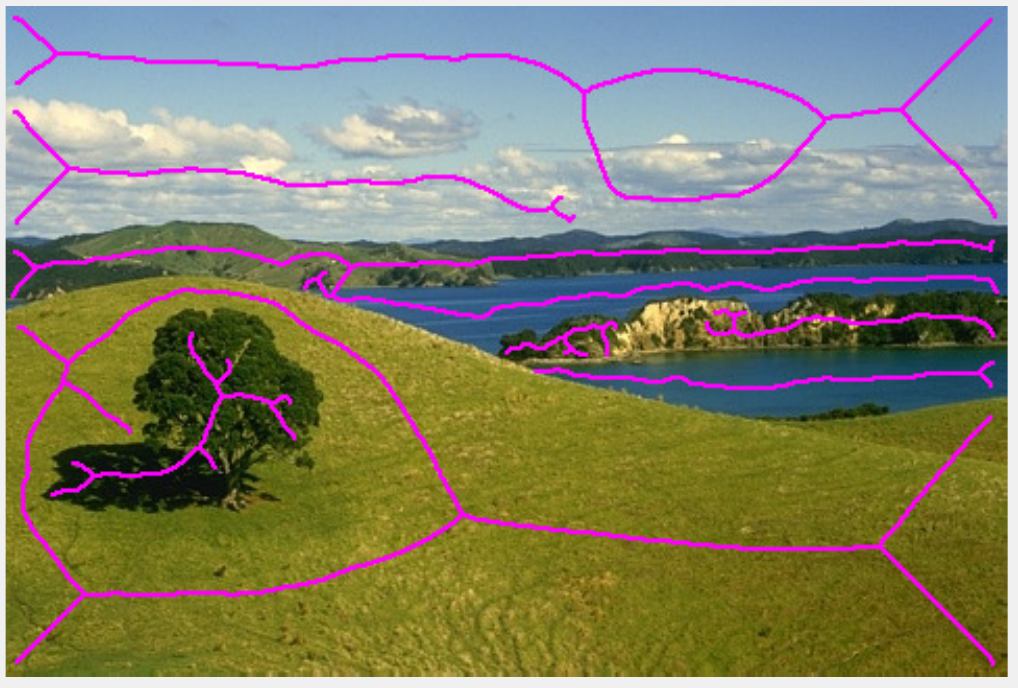}} &
		\fbox{\includegraphics[width=0.31\textwidth]{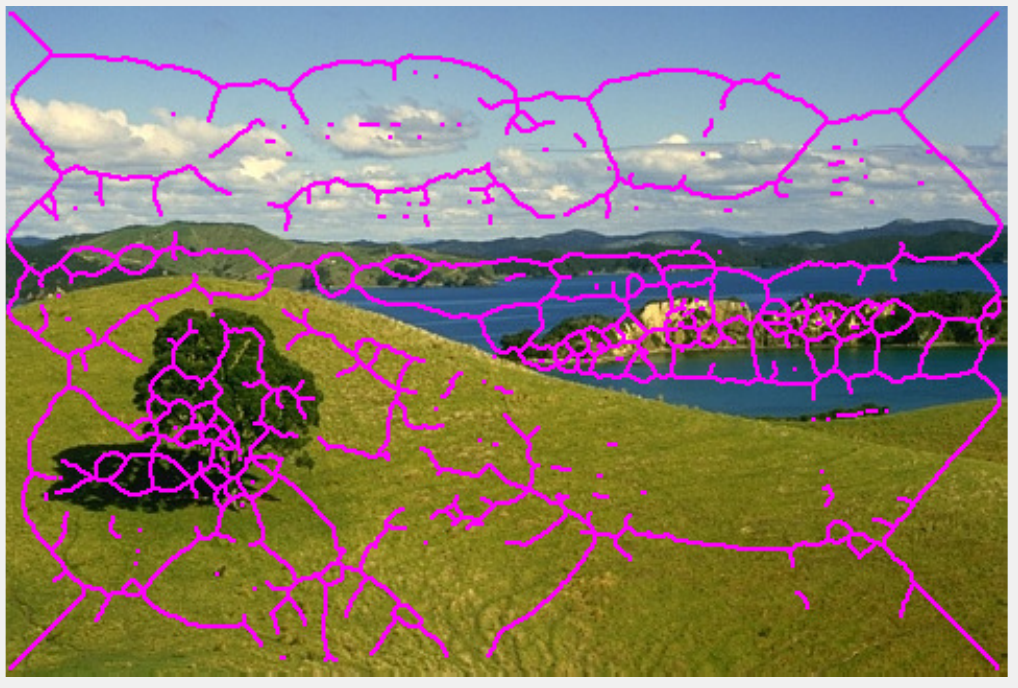}}\\
	\end{tabular}
	\caption{Qualitative results. \textbf{Left to right:} Ground-truth (1 annotation), ASG result, AMAT result. }
	\label{fig:qualitatives3}
\end{figure*}

\begin{figure*}[!hbt]
	\begin{tabular}{@{\hskip3pt}c@{\hskip3pt} c @{\hskip3pt}c@{\hskip3pt}}
				\fbox{\includegraphics[width=0.31\textwidth]{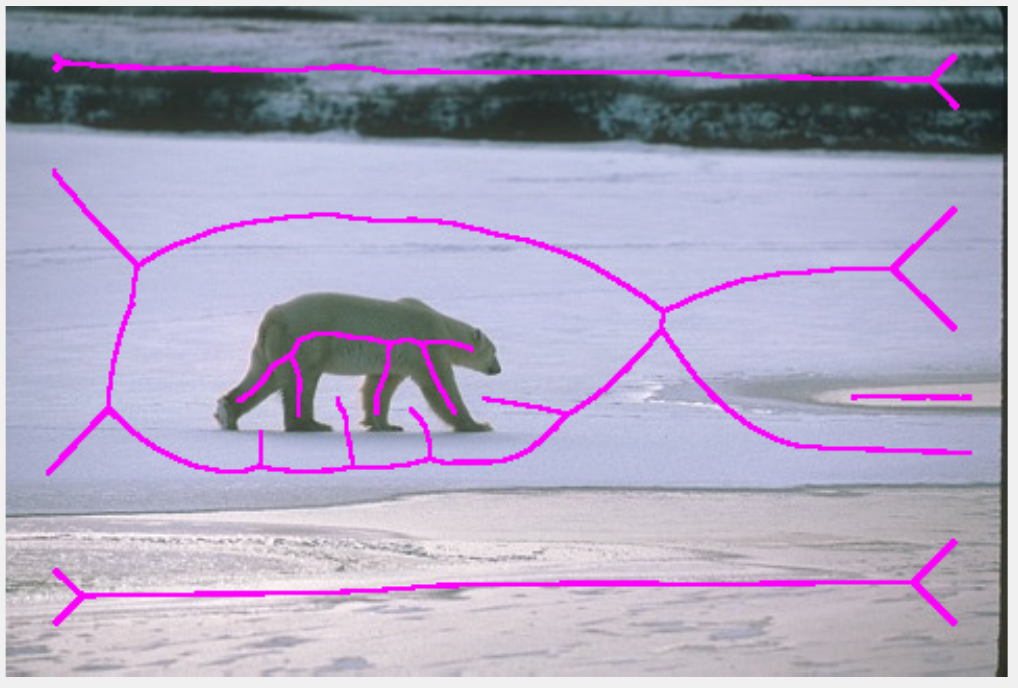}} &
		\fbox{\includegraphics[width=0.31\textwidth]{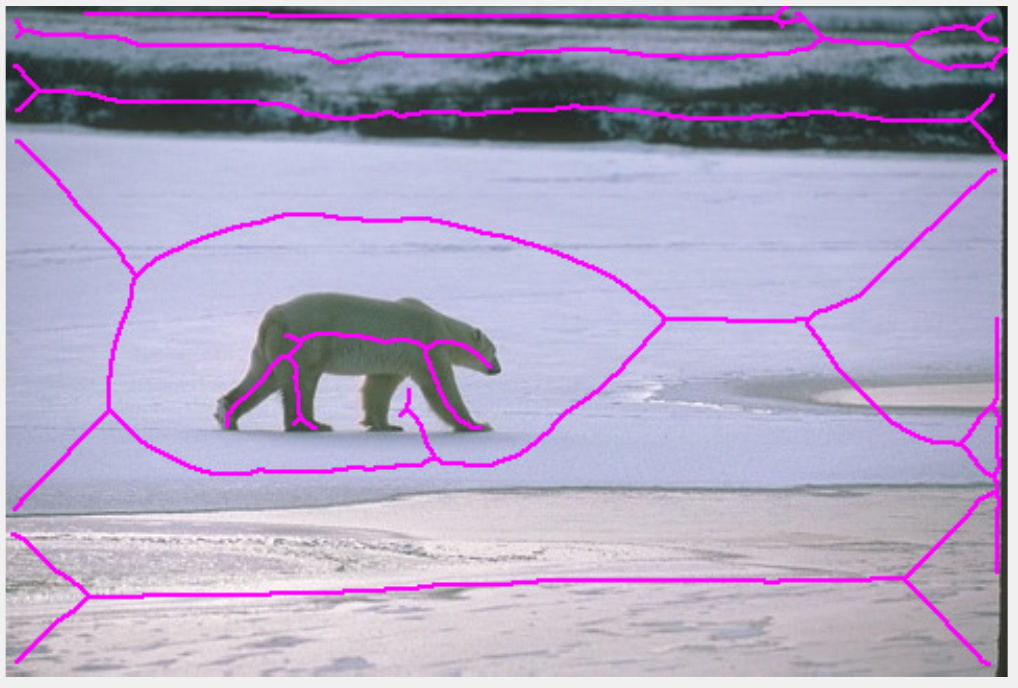}} &
		\fbox{\includegraphics[width=0.31\textwidth]{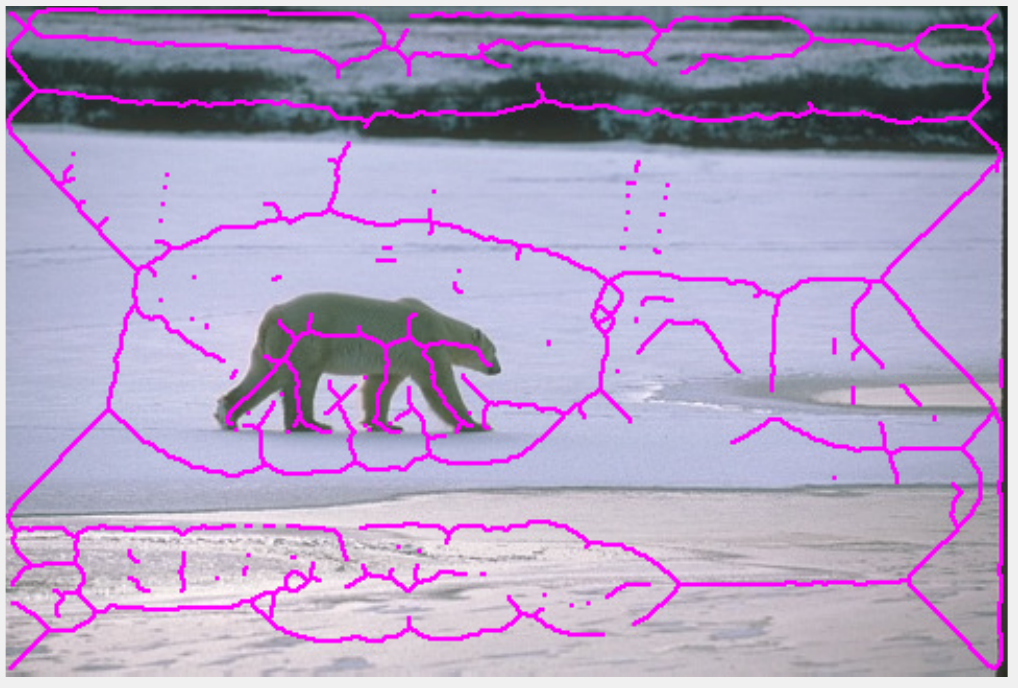}}\\
        \fbox{\includegraphics[width=0.31\textwidth]{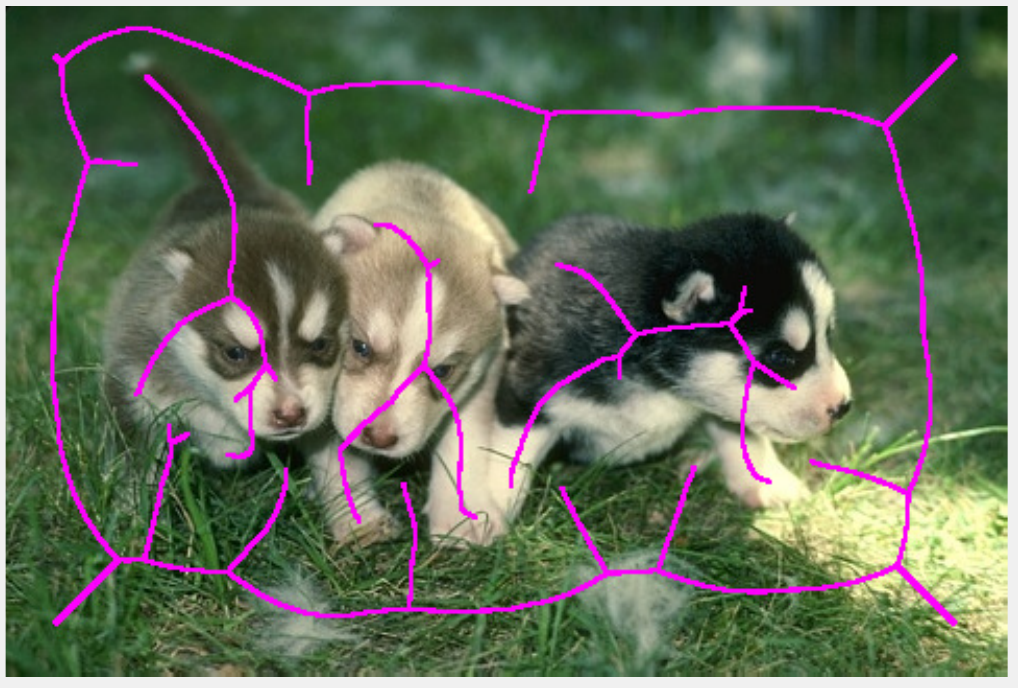}} &
		\fbox{\includegraphics[width=0.31\textwidth]{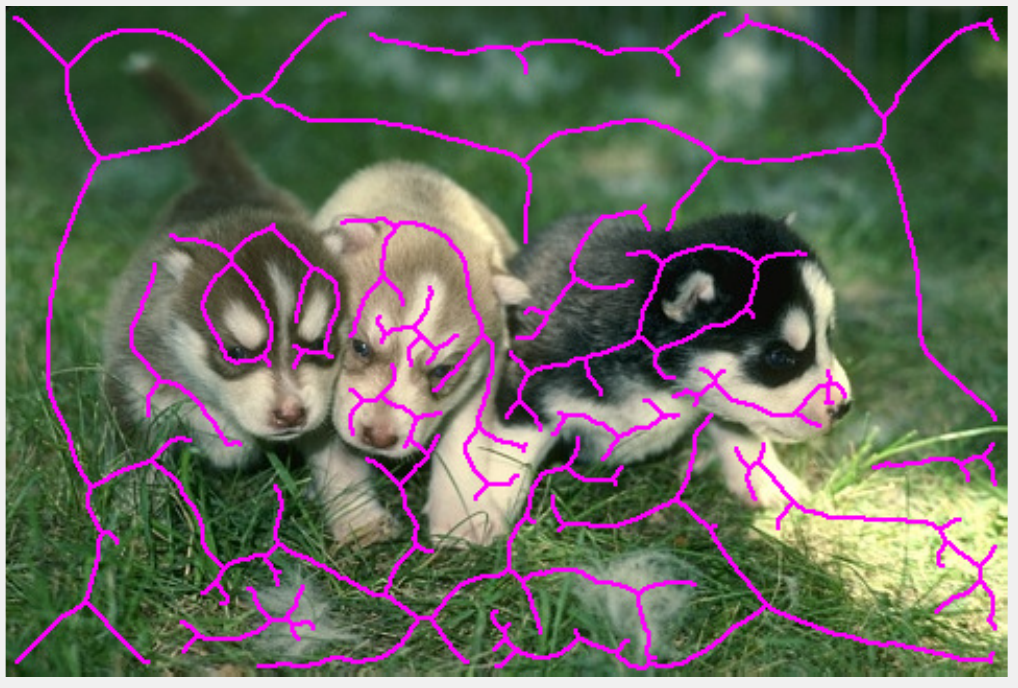}} &
		\fbox{\includegraphics[width=0.31\textwidth]{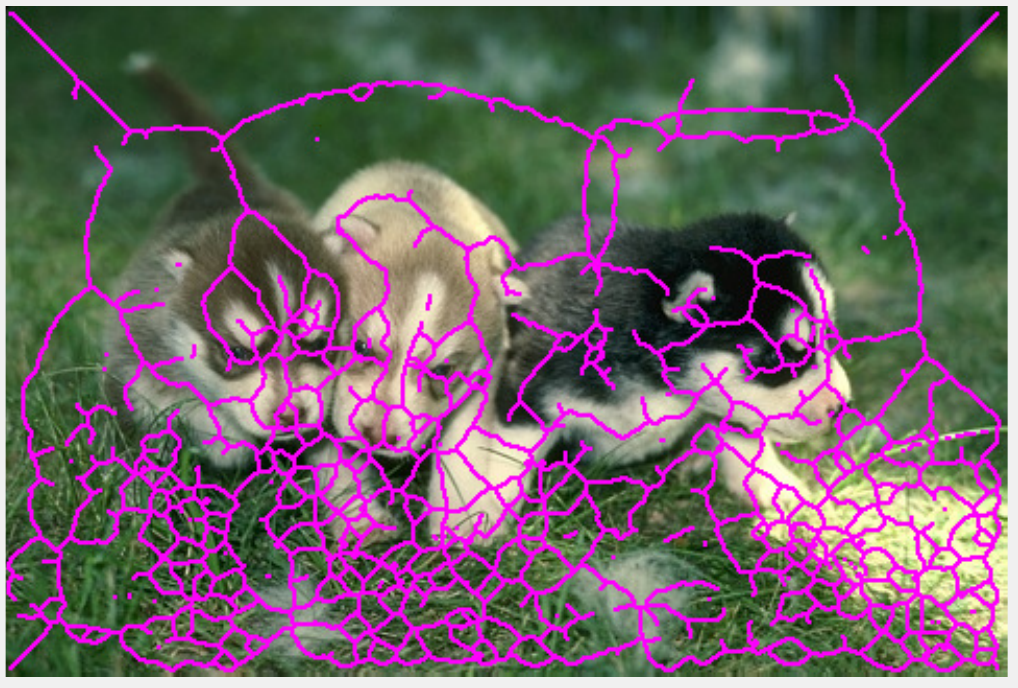}}\\
		\fbox{\includegraphics[width=0.31\textwidth]{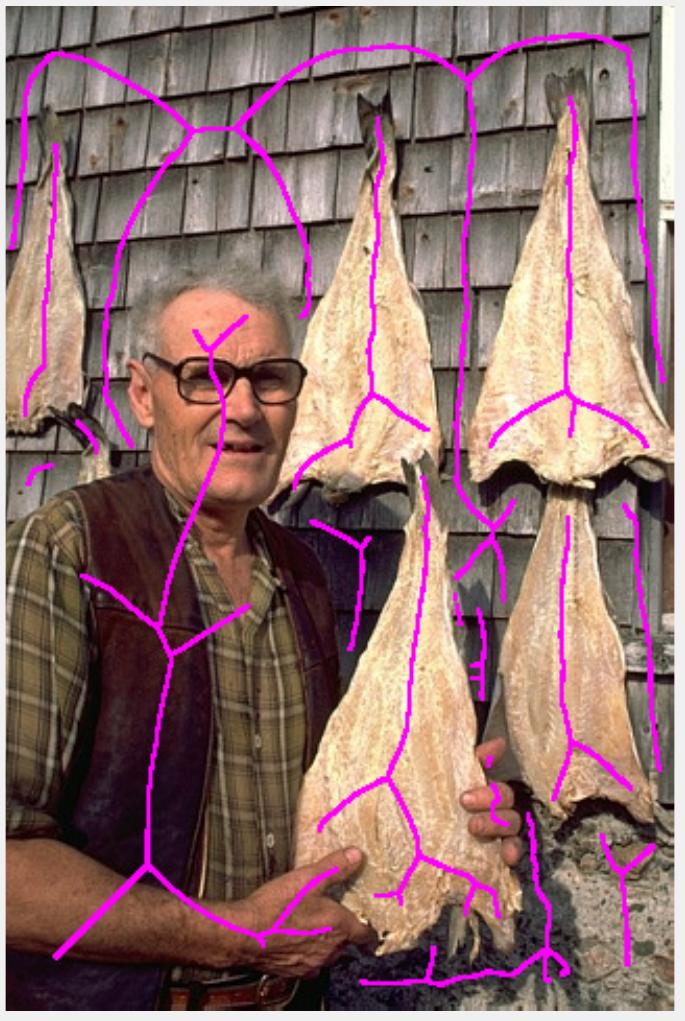}} &
		\fbox{\includegraphics[width=0.31\textwidth]{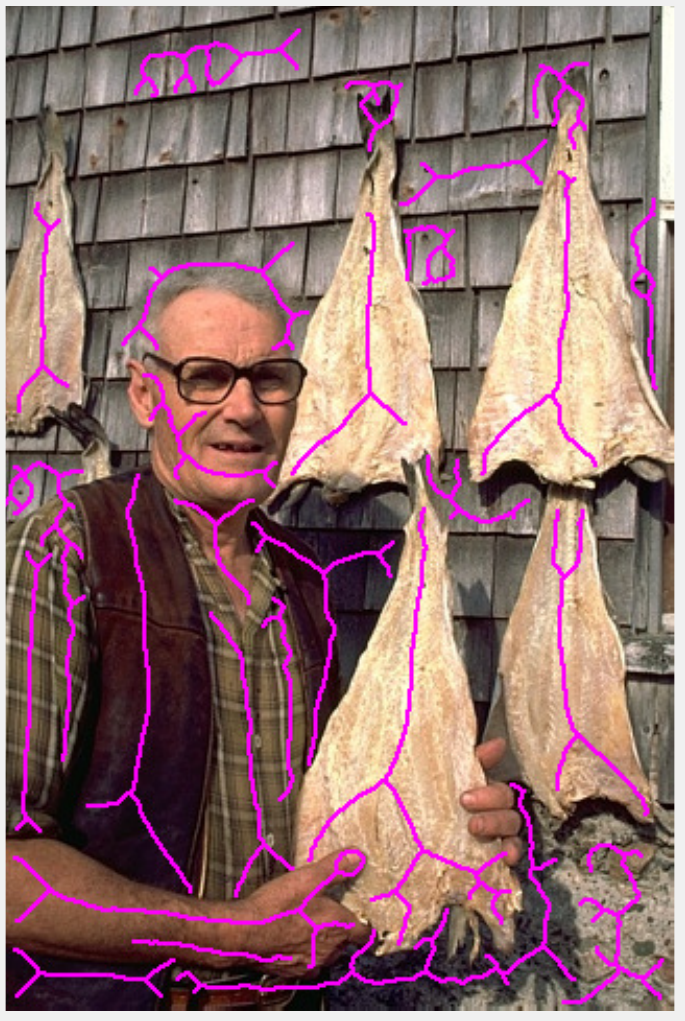}} &
		\fbox{\includegraphics[width=0.31\textwidth]{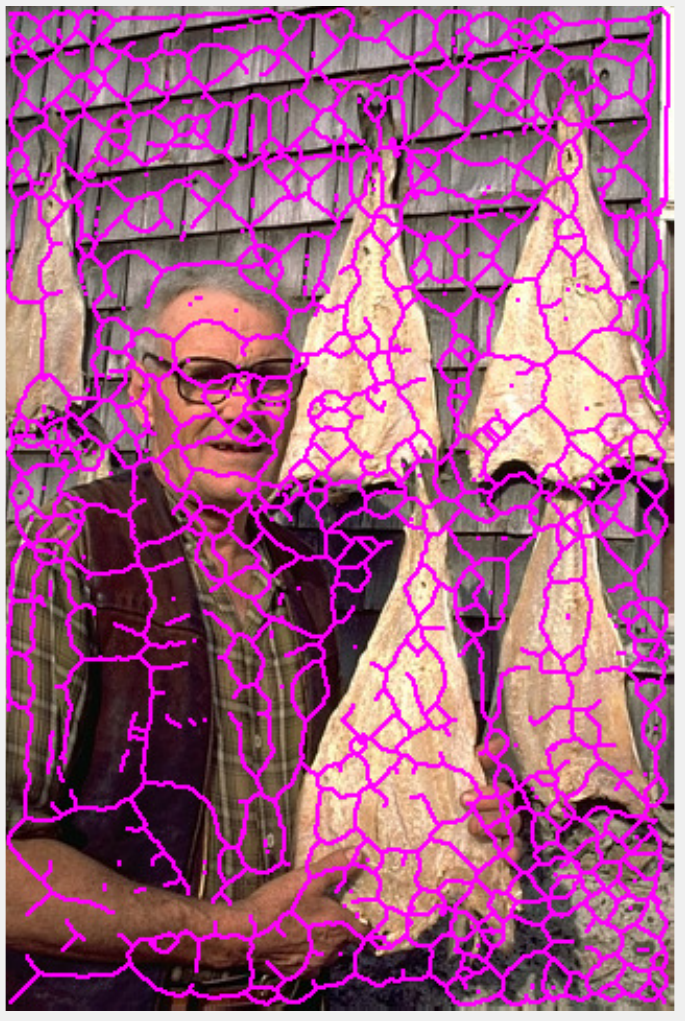}}\\
	\end{tabular}
	\caption{Qualitative results. \textbf{Left to right:} Ground-truth (1 annotation), ASG result, AMAT result. }
	\label{fig:qualitatives4}
\end{figure*}